\pdfoutput=1
\documentclass[11pt]{article}
\usepackage{multirow}
\usepackage{graphicx}
\usepackage[preprint]{acl}
\usepackage{booktabs}
\usepackage{siunitx}
\usepackage{graphicx}
\usepackage{times}
\usepackage{latexsym}
\usepackage{enumitem}
\usepackage{array}
\usepackage{lscape}
\usepackage{algorithm}
\usepackage{algorithmic}
\usepackage{booktabs}
\usepackage{float}
\usepackage{geometry} % for adjusting page layout
\usepackage{caption} % for customizing captions
\usepackage{siunitx}  % for aligning numbers in columns
\usepackage[T1]{fontenc}
\usepackage[utf8]{inputenc}

\usepackage{microtype}

\usepackage{inconsolata}

\usepackage{graphicx}
\usepackage{svg}

\title{MAPWise: Evaluating Vision-Language Models for Advanced Map Queries}

\author{
    \textbf{Srija Mukhopadhyay\textsuperscript{\ddag}}, 
    \textbf{Abhishek Rajgaria\textsuperscript{\dag}},
    \textbf{Prerana Khatiwada}\textsuperscript{\#}\\
    \textbf{Vivek Gupta\thanks{Corresponding author, work done at UPenn.}},
    \textbf{Dan Roth}\textsuperscript{§} \\
    \textsuperscript{\ddag}IIIT Hyderabad,
    \textsuperscript{\dag}University of Utah,
    \textsuperscript{\#}University of Delaware,\\
    \textsuperscript{*}Arizona State University,
    \textsuperscript{§}University of Pennsylvania\\
    \texttt{\small srija.mukhopadhyay@research.iiit.ac.in}, 
    \texttt{\small abhishekemail@domain.com},
    \texttt{\small preranak@udel.edu}, \\
    \texttt{\small vgupt140@asu.edu},
    \texttt{\small danroth@seas.upenn.edu}\
}

\begin{document}
\maketitle
\begin{abstract}
Vision-language models (VLMs) excel at tasks requiring joint understanding of visual and linguistic information. A particularly promising yet under-explored application for these models lies in answering questions based on various kinds of maps. This study investigates the efficacy of VLMs in answering questions based on choropleth maps, which are widely used for data analysis and representation. To facilitate and encourage research in this area, we introduce a novel map-based question-answering benchmark, consisting of maps from three geographical regions (United States, India, China), each containing 1000 questions. Our benchmark incorporates 43 diverse question templates, requiring nuanced understanding of relative spatial relationships, intricate map features, and complex reasoning. It also includes maps with discrete and continuous values, encompassing variations in color-mapping, category ordering, and stylistic patterns, enabling comprehensive analysis. We evaluate the performance of multiple VLMs on this benchmark, highlighting gaps in their abilities and providing insights for improving such models. 
\looseness -1
\end{abstract}

\section{Introduction}
\begin{figure}[ht]
  \centering
  \includegraphics[width=0.75\linewidth]{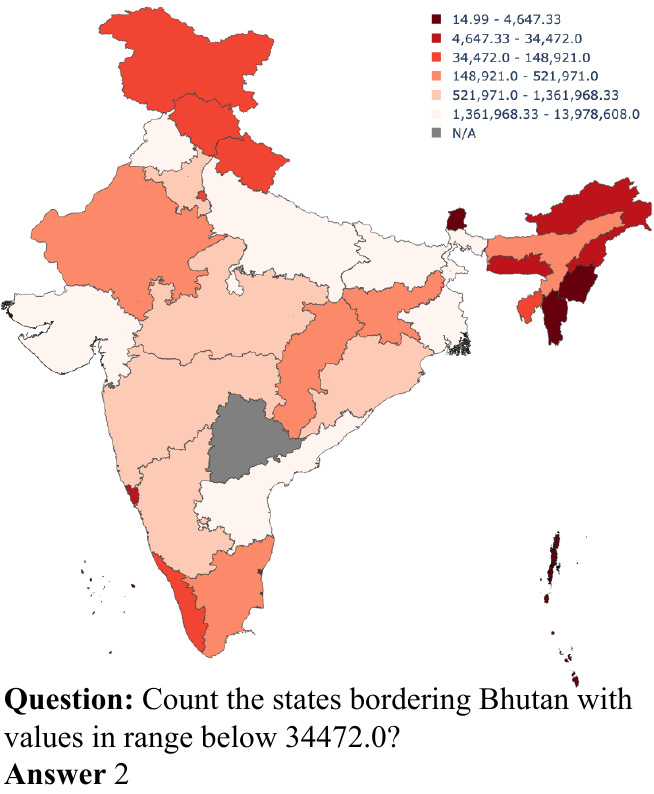}
% \captionsetup{justification=centering}  
\vspace{-0.5em}
  \caption{A question-map pair from our MAPWise dataset and the corresponding gold truth answer.}
  \label{fig:india_map}
  \vspace{-0.5em}
\end{figure}
Vision-Language Models (VLMs) have demonstrated impressive capabilities in tasks requiring joint understanding of visual information and natural language.  They have achieved significant success in areas like visual question answering (VQA) \cite{salaberria2023image, chaudhry2020leaf}, image generation \cite{zhao2024bridging}, and multimodal sentiment analysis \cite{yi2024vlp2msa}.  However, when applied to map-based question answering, the reasoning abilities of these models remain largely unexplored \cite{chang2022mapqa}.
% \looseness -1
% Recent research in Visual-Question Answering (VQA) \cite{kafle2018dvqa,kahou2017figureqa} has primarily focused on photographs and human-generated charts, such as bar, line, and pie charts (known as ChartQA \cite{masry2022chartqa}).  However, the understanding of maps, particularly choropleth maps, has been greatly under-explored.  

Choropleth maps, which use varying shades or colors to represent geographical data, present a unique challenge \cite{chang2022mapqa}.  While humans can readily grasp the spatial patterns and information conveyed by these color variations, their interpretation poses a significant challenge for visual language models and other analytical tools. This difficulty arises from the inherent challenge of translating visual data represented by different colors or shades into simpler, tabular formats.

This research addresses this gap by analyzing the performance of VLMs in answering questions related to choropleth maps representing different geographical regions (Figure \ref{fig:india_map}).
% \looseness -1
We aim to answer the following research questions:  

$(RQ1)$ How effectively can VLMs answer questions about Choropleth maps of different geographical regions? 

$(RQ2)$ What prompting strategies can improve the performance of models for Map Visual Question Answering (Map-VQA)? 

$(RQ3)$ What biases are present in these models with regards to Map VQA?

$(RQ4)$ How effectively do these models attend to the provided map when performing visual question-answering tasks?

To address these research questions, we created a novel dataset, \textbf{MAPWise}, specifically for map-based VQA. This dataset comprises \textbf{1,000} questions for three geographical regions: the United States, India, and China. The questions were manually created based on \textbf{43} unique templates, designed to evaluate model capabilities across a diverse range of topics, from data extraction to complex reasoning.

Furthermore, the dataset includes various map representations, including maps with and without annotations, a diverse range of colormaps, and stylistic patterns like hatching, creating a robust benchmark. We have used this dataset for experimentation across various leading VLMs and MMLMs, using diverse prompting techniques to establish a viable baseline. Our study also included an analysis of model performance on counterfactual maps. These maps featured imaginary state names, jumbled state names, and counterfactual statistics. Our analysis aimed to not only understand how well the models relied on the provided map data but also to what extent they relied on their internal knowledge. The contributions of our study are threefold:
\begin{itemize}[noitemsep]
    \item \textbf{\textit{Dataset}:} The MAPWise dataset, tailored for choropleth maps, provides diverse questions that test various aspects of geographical and spatial understanding.
    \item \textbf{\textit{Models}:} Baseline performances using VLMs provide a reference point for future research in map-based VQA. We also included human baseline scores for a more comprehensive analysis. 
    \item \textbf{\textit{Bias and Counterfactual Analysis:}} In-depth analysis of biases present in the models along with our counterfactual analysis highlights areas of struggle and offers insights for improvement. 
\end{itemize}

\section{ The MAPWise Dataset} 
This section details the creation process of the MAPWise dataset, including data gathering, manual question creation, and dataset validation. 

\subsection{Dataset Creation}
\paragraph{Data Sources.}
The \textbf{MAPWise} dataset was created using data from three countries: India, USA, and China. We have meticulously chosen reliable sources to gather socioeconomic and demographic statistics for each country, as described below.

i) For \textbf{India}, we sourced data from the Reserve Bank of India's \textit{"Handbook of Statistics on Indian States."} This resource provides extensive data across various periods, including details such as state-wise cold storage capacity, rural population figures, and the area of non-food grains like cotton.

ii) For \textbf{USA}, the primary data source was the \textit{"Kaiser Family Foundation"}, which specializes in healthcare statistics. This includes information on health insurance coverage for adults without dependent children, age-adjusted suicide rates, and weekly COVID-19 vaccine allocations.

iii) For \textbf{China}, we obtained data from the \textit{"National Bureau of Statistics of China."} This source provides data such as household consumption expenditure, urban unemployment rates and natural growth rate. 

\begin{figure*}[ht]
  \centering
  \includegraphics[width=\linewidth]{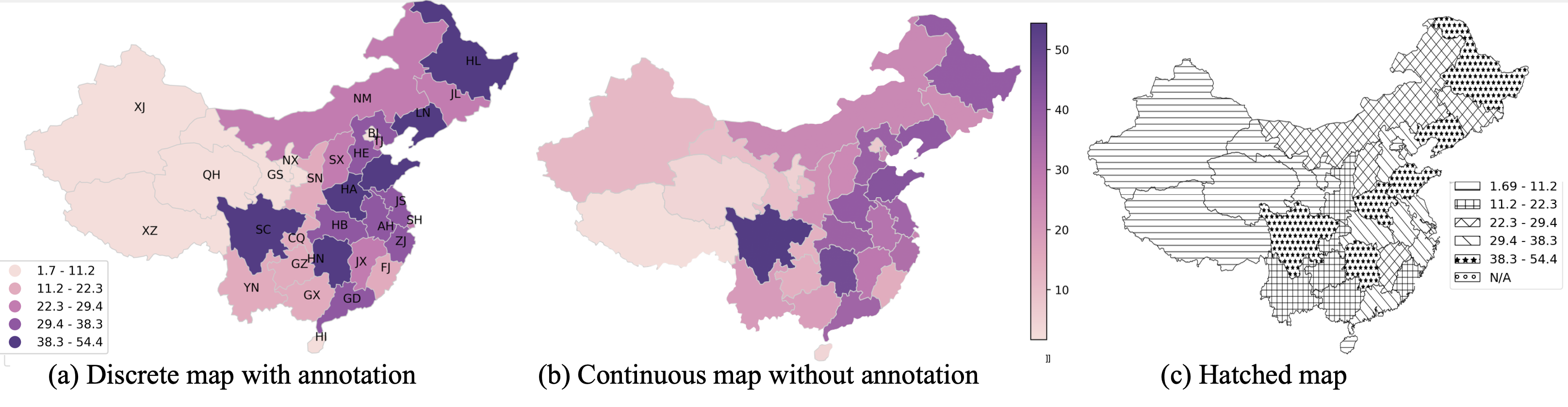}
  % \captionsetup{justification=centering}
  \caption{\small Examples of map with annotations, without annotations for the same underlying data. Additionally, hatched maps were created to better assess model understanding and performance.}
  \label{fig:annotated_maps}
\end{figure*}

\paragraph{Map Variations.} The dataset consists of maps representing data in two primary forms: discrete, where the legend is divided into distinct groups and continuous, where the legend is distributed over a spectrum. The maps also include variations in the presence or absence of annotations, which provide additional contextual information. Our dataset also includes maps with black-and-white textured patterns or hatches for discrete data, different color-map variations (light, dark, and gradient scales), and varying paper background colors (white and grey). These variations test the models' capability to handle diverse visual presentations. We generated maps with annotations, without annotations, and with hatching for each country using the Plotly library.

\paragraph{Question Generation.} To create a comprehensive and insightful benchmark, we designed question templates with varying levels of difficulty, ranging from simple \textit{yes/no} questions to more complex \textit{region association} questions that required reasoning based on relative locations. 

The dataset includes three major question types: Binary questions, which require a simple yes or no answer based on the map; Direct Value Extraction questions, which ask for a specific numerical or nominal value related to a particular region or the legend; and Region Association questions, which involve identifying or counting regions meeting some specific criteria, often requiring geospatial reasoning and reasoning about relative regions. 

Each question could have answers in one of the following formats: Binary (Yes/No), Single Word, Count, List, Range, and Ranking. Examples of these are shown in Table \ref{questions}. All questions were manually created by expert annotators, with the help of provided templates, with 10 questions created for each map. Overall, we created 1000 question-answer pairs for each country. The statistics of our final dataset have been summarized in Table \ref{tab:dataset-stat}. 

\begin{table}[ht!]
\centering
\resizebox{\columnwidth}{!}{%
\begin{tabular}{@{}ll@{}}
\toprule
\textbf{Answer Type} & \multicolumn{1}{c}{\textbf{Example Question}} \\ \midrule
Binary & \begin{tabular}[c]{@{}l@{}}Yes or no: California is an outlier compared to its \\ neighbours?\end{tabular} \\ \midrule
Single Word & \begin{tabular}[c]{@{}l@{}}Name the eastern-most state that belongs to a higher \\ value range compared to all its neighbours.\end{tabular} \\ \midrule
List & \begin{tabular}[c]{@{}l@{}}Which states in the East China Sea region have a value\\ higher than state Guangdong?\end{tabular} \\ \midrule
Range & What is the least value range in the west coast region? \\ \midrule
Count & \begin{tabular}[c]{@{}l@{}}How many states bordering Canada have a value lower\\ than New Mexico?\end{tabular} \\ \midrule
Ranking & \begin{tabular}[c]{@{}l@{}}Rank Rajasthan, Gujarat and Jammu and Kashmir in \\ terms of the legend value in region bordering Pakistan.\end{tabular} \\ \bottomrule
\end{tabular}%
}
\caption{Example questions along with the different types of possible answers.}
\label{questions}
\end{table}

\paragraph{Dataset Validation.}
The generated questions were initially validated by expert annotators (detailed in Appendix \ref{sec:dataset_validation_prc}). Following that, we carried out a process of human evaluation that played a critical role in confirming the accuracy of our dataset. It also served as a benchmark for comparisong model performance. Table \ref{tab:human_baseline_results} presents human evaluation metrics for the three countries. 

\begin{table}[ht]
\centering
\footnotesize
\begin{tabular}{@{}l|lccc@{}}
\toprule
\textbf{Type} & \textbf{Country} & \textbf{USA} & \textbf{India} & \textbf{China} \\ \midrule
% \multicolumn{4}{c}{\textbf{General Statistics}} \\ \midrule
& Total & 97 & 100 & 100 \\ 
% \multicolumn{4}{c}{\textbf{Map Types}} \\ \midrule
\textbf{Maps} & Continuous & 33 & 51 & 49 \\
& Discrete & 64 & 49 & 51 \\ \midrule
% \multicolumn{4}{c}{\textbf{Question types}} \\ \midrule
% \multicolumn{4}{c}{\textbf{Answer types}} \\ \midrule
& Binary & 449 & 456 & 441 \\
& Single Word & 235 & 196 & 187 \\
\textbf{Answer} &List & 137 & 153 & 163 \\
\textbf{Types} & Range & 130 & 103 & 112 \\
& Count & 49 & 95 & 97 \\
& Ranking & 0 & 29 & 26 \\ \midrule
\textbf{Question} & Relative regions & 145 & 206 & 214 \\ 
\bottomrule
\end{tabular}
\caption{Overview of MAPWise statistics.}
\vspace{-2.0em}
\label{tab:dataset-stat}
\end{table}

\section{Experimental Evaluation}
This section outlines our experimental setup: we selected a mix of closed-source and open-source Vision-Language Models (VLMs) and Multimodal Large Language Models (MLLMs) for a comprehensive analysis. These models were tested with various prompting techniques, and we developed an evaluation metric to assess different answer types.

\subsection{Baseline Models}
\paragraph{\textit{Closed-Source MLLMs.}} For analysis on closed source models, we used Gemini 1.5 Flash \cite{geminiteam2024gemini} and GPT-4o \cite{openai2024gpt4}. These models are known for their advanced features and proprietary implementations. 

\paragraph{\textit{Open-Source VLMs.}} We selected CogAgent, InternLM XComposer 2, Idefics 2, and Qwen VL.  CogAgent-VQA \cite{hong2023cogagent} is an 18-billion-parameter VLM specializing in GUI understanding and navigation. InternLM-XComposer2 \cite{dong2024internlmxcomposer2}, an adaptation of InternLM2-7B \cite{cai2024internlm2}, excels in producing high-quality long-text multimodal content and reasoning within visual-language understanding contexts. QwenVL \cite{bai2023qwenvl}, a generalist 7-billion-parameter VLM built on top of Qwen-LM \cite{bai2023qwen}, leverages adapted visual encoders and general and multi-task pretraining. These models were chosen due to their accessibility and contributions to the research community, each offering distinct approaches to processing and interpreting visual information. 

\subsection{Prompting Strategies}
We evaluated the baseline models under two distinct prompting settings:

\begin{enumerate}
    \item \textbf{\textit{Zero-Shot Chain-of-Thought Prompting (COT)}}.
   We leverage the Chain-of-Thought \cite{wei2023chainofthought} prompting, presenting the VLM with a map and a question, prompting it to reason through the steps leading to its final answer. 
    \item \textbf{\textit{Explicit Extraction and Reasoning (EER)}}. Here, we created a custom prompt that explicitly outlined the reasoning steps the model should follow to answer the specific question.  This prompt was broken down into four distinct reasoning steps:
    
    - \textit{Extraction of Regions}. The model was prompted to identify the regions whose data was required to answer the question. 
    
    - \textit{Extraction of Relevant Places}. Next, the model was instructed to extract the specific locations or places associated with the identified regions.
    
    - \textit{Extraction of Values from Legend}. The model was then directed to extract the values corresponding to those regions from the map's legend. 
    
    - \textit{Reasoning based on Extracted Values}. Finally, the model was prompted to reason based on the extracted values to arrive at the final answer. 
    
This approach helped break down the reasoning process into smaller, more manageable steps, preventing the model from becoming overwhelmed and guiding it towards a more focused and structured reasoning process.
\end{enumerate}

During the evaluation, all models were given the same prompt in order to fairly and consistently assess their ability to reason. The prompts used have been presented in the \nameref{appendix}.

\subsection{Evaluation Details}
The evaluation process adapts to various answer types within in the dataset by employing tailored metrics and criteria for each specific answer type. Additionally, normalization was applied wherever necessary to ensure consistency and accuracy in the assessment.

For binary \textit{yes/no} and integer \textit{count} answers, we implemented an exact match criterion and accuracy as the evaluation metric. For single-word answers, as some questions have multiple applicable responses, we employed the recall metric for better evaluation. For state names, a valid answer could be either a two-digit state code or the full state name. For ranges, we first normalized the ranges to absolute values (e.g. \textit{1k to 1000}) and then compared them. For discrete maps, only exact match was expected, whereas for continuous maps, we gave a full score of 1 for exact match and a partial score of 0.5 for overlapping responses.

For list type answer, we used precision and recall metrics because predicted lists often contained irrelevant states (\textit{false positives}) and missed relevant states (\textit{false negatives}).

For rank-type answers, we prompted the model to assign ranks to states based on map values. However, due to the difficulty in accurately distinguishing shades, models frequently assigned states to wrong shades, resulting in multiple states sharing the same rank despite differing shades. Additionally, for some questions, ground truth involved multiple states in the same rank because of states having identical shades or patterns. To evaluate this, we designed a “Rank-wise Precision (RWP)” method, computing precision for each rank and then averaging across all ranks. We also evaluated other ranking metrics, including Mean Reciprocal Rank (MRR) and Mean Average Precision (MAP), as detailed in Appendix \ref{sec:rwp_explanation}. 

\paragraph{\textit{Note for Open Source VLMs.}} Smaller models, like QwenVL, CogAgent, and InternLM, faced challenges in producing answers in the desired format. To address this, we used an \textbf{"LLM as an Extractor"} approach, using Gemini 1.5 Flash to extract answers from their outputs.  Manual verification of \textit{150} samples confirmed that Gemini primarily acted as a extracting and formatting tool, preserving the original model's answer in \textit{138} cases. In the remaining \textit{12} cases, the original model had not clearly answered the question, for which Gemini reported \textit{"Answer cannot be extracted".}

\section{Results and Analysis}
\paragraph{MAPWise: A Challenging Benchmark.} The MAPWise dataset presents a compelling benchmark for evaluating the reasoning abilities of current Vision-Language Models (VLMs). As shown in Table \ref{tab:model-perf}, models consistently perform significantly worse than the human baseline, particularly with questions requiring intricate reasoning, such as counting or providing a list of regions where the difference in scores is close to \textbf{50\%} on average. This substantial performance gap highlights a significant limitation in the reasoning capabilities of existing VLMs, underscoring the need for further research to bridge this gap. 

\paragraph{Model Performance Comparison.} While model performance varied across different answer types and countries, GPT-4o consistently emerged as the top performer in most categories, closely followed by Gemini 1.5 Flash (as shown in Table \ref{tab:model-perf}). Notably, Gemini demonstrated superior performance on hatched maps (as seen in Table \ref{tab:gpt_vs_gem}), likely due to its stronger legend resolution and data extraction capabilities.  However, GPT-4o's  robust reasoning skills generally led to better scores across other task types.

\begin{table}[ht!]
\resizebox{\columnwidth}{!}{%
\begin{tabular}{c|ccccccc}
\hline
\textbf{Model}          & \textbf{\begin{tabular}[c]{@{}c@{}}Binary \\ Acc\end{tabular}} & \textbf{\begin{tabular}[c]{@{}c@{}}Single \\ Recall \end{tabular}} & \textbf{\begin{tabular}[c]{@{}c@{}}Count \\ Acc\end{tabular}} & \textbf{\begin{tabular}[c]{@{}c@{}}Range \\ Acc\end{tabular}} & \textbf{\begin{tabular}[c]{@{}c@{}}List \\ Precision\end{tabular}} & \textbf{\begin{tabular}[c]{@{}c@{}}List \\ Recall\end{tabular}} & \textbf{\begin{tabular}[c]{@{}c@{}}Rank\\ RWP\end{tabular}} \\ \hline
\textit{\textbf{Human}} & 96.97           & 86.21           & 80.00          & 89.29          & 98.61                                                              & 94.44                                                           & 91.67                                                       \\ \hline
\textbf{GPT-4o}         & \textbf{71.52}  & 40.06           & \textbf{35.48} & \textbf{55.75} & \textbf{49.94}                                                     & \textbf{49.17}                                                  & \textbf{54.94}                                              \\
\textbf{Gemini}         & 56.36           & 38.49           & 24.47          & 40.27          & 34.55                                                              & 45.11                                                           & 38.69                                                       \\
\textbf{Intern-LM}      & 56.80           & 32.37           & 17.02          & 13.27          & 20.14                                                              & 24.02                                                           & 35.71                                                       \\
\textbf{Idefics}        & 54.71           & \textbf{43.59}  & 13.83          & 28.76          & 32.19                                                              & 38.29                                                           & 45.24                                                       \\
\textbf{CogAgent}       & 43.27           & 25.32           & 9.57           & 16.81          & 19.62                                                              & 26.32                                                           & 41.36                                                       \\
\textbf{QwenVL}         & 37.75           & 22.33           & 4.26           & 6.64           & 17.00                                                              & 23.60                                                           & 17.31                  \\
\hline
\end{tabular}%
}
\caption{Table with results for different models when evaluated on annotated maps of India using the zero-shot COT prompt, compared against the human baseline. Here "Acc" stands for Accuracy.}
\vspace{-1.0em}
\label{tab:model-perf}
\end{table}

\begin{table}[ht!]
\resizebox{\columnwidth}{!}{%
\begin{tabular}{cccccccc}
\hline
\textbf{Model}          & \textbf{\begin{tabular}[c]{@{}c@{}}Binary \\ Acc\end{tabular}} & \textbf{\begin{tabular}[c]{@{}c@{}}Single \\ Recall \end{tabular}} & \textbf{\begin{tabular}[c]{@{}c@{}}Count \\ Acc\end{tabular}} & \textbf{\begin{tabular}[c]{@{}c@{}}Range \\ Acc\end{tabular}} & \textbf{\begin{tabular}[c]{@{}c@{}}List \\ Precision\end{tabular}} & \textbf{\begin{tabular}[c]{@{}c@{}}List \\ Recall\end{tabular}} & \textbf{\begin{tabular}[c]{@{}c@{}}Rank\\ RWP\end{tabular}} \\ \hline
\multicolumn{8}{c}{\textbf{USA}}                                                                                                                                                                                                                                                             \\ \hline
\textbf{Gemini}         & 49.36           & 56.20           & 51.22          & 53.95          & 20.51                                                         & 35.35                                                          & -                                                           \\
\textbf{GPT}            & 49.78           & 16.14           & 26.83          & 26.67          & 26.96                                                         & 32.60                                                          & -                                                           \\ \hline
\multicolumn{8}{c}{\textbf{India}}                                                                                                                                                                                                                                                           \\ \hline
\textbf{Gemini}         & 52.75           & 48.65           & 23.53          & 34.38          & 38.95                                                         & 47.33                                                          & 38.89                                                       \\
\textbf{GPT}            & 49.72           & 28.38           & 31.37          & 30.77          & 34.19                                                         & 36.27                                                          & 53.57                                                       \\ \hline
\multicolumn{8}{c}{\textbf{China}}                                                                                                                                                                                                                                                           \\ \hline
\textbf{Gemini}         & 53.80           & 55.41           & 22.22          & 40.32          & 29.61                                                         & 45.41                                                          & 60.42                                                       \\
\textbf{GPT}            & 45.11           & 20.56           & 27.78          & 18.03          & 33.33                                                         & 34.40                                                          & 39.58                                                       \\ \hline
\end{tabular}%
}
\caption{Table showing scores for Gemini 1.5 and GPT-4o for hatched maps using zero shot COT prompt. Here, "Acc" stands for Accuracy.}
\vspace{-1.0em}
\label{tab:gpt_vs_gem}
\end{table}

While open-source models generally lag behind their closed-source counterparts in performance, Idefics and InternLM demonstrate surprisingly strong results. However, we observed that open-source models struggle significantly with questions requiring complex reasoning, with QwenVL achieving a low 4.26\% accuracy on tasks involving counting.  This stark difference underscores the crucial need for models not only to excel in data extraction but also to possess sophisticated reasoning skills, particularly in the domain of geo-spatial reasoning.

\begin{table}[ht]
\resizebox{\columnwidth}{!}{%
\begin{tabular}{cccccccc}
\hline
\textbf{Prompt}          & \textbf{\begin{tabular}[c]{@{}c@{}}Binary \\ Acc\end{tabular}} & \textbf{\begin{tabular}[c]{@{}c@{}}Single \\ Recall \end{tabular}} & \textbf{\begin{tabular}[c]{@{}c@{}}Count \\ Acc\end{tabular}} & \textbf{\begin{tabular}[c]{@{}c@{}}Range \\ Acc\end{tabular}} & \textbf{\begin{tabular}[c]{@{}c@{}}List \\ Precision\end{tabular}} & \textbf{\begin{tabular}[c]{@{}c@{}}List \\ Recall\end{tabular}} & \textbf{\begin{tabular}[c]{@{}c@{}}Rank\\ RWP\end{tabular}} \\ \hline
\multicolumn{8}{c}{\textbf{GPT-4o}}                                                                                                                                                                                                                                                                                                                                                                                       \\ \hline
\textbf{COT}    & 66.97                                                & 47.53                                                    & 50.52                                                & 59.40                                                & 53.93                                                     & 57.56                                                 & 46.58                                               \\
\textbf{EER}    & 63.33                                                & 60.65                                                    & 45.36                                                & 59.83                                                & 43.47                                                     & 46.48                                                 & 56.62                                               \\ \hline
\multicolumn{8}{c}{\textbf{Gemini 1.5 Flash}}                                                                                                                                                                                                                                                                                                                                                                             \\ \hline
\textbf{COT}    & 62.27                                                & 51.83                                                    & 13.40                                                & 52.97                                                & 22.76                                                     & 38.96                                                 & 53.63                                               \\
\textbf{EER}    & 61.50                                                & 54.09                                                    & 24.74                                                & 52.14                                                & 23.01                                                     & 39.54                                                 & 49.54                                               \\ \hline
\multicolumn{8}{c}{\textbf{InternLM-XComposer2}}                                                                                                                                                                                                                                                                                                                                                                          \\ \hline
\textbf{COT}    & 54.09                                                & 50.54                                                    & 21.65                                                & 34.32                                                & 21.67                                                     & 29.91                                                 & 46.15                                               \\
\textbf{EER}    & 53.86                                                & 29.25                                                    & 18.56                                                & 28.39                                                & 26.63                                                     & 39.78                                                 & 28.21                                               \\ \hline
\multicolumn{8}{c}{\textbf{Idefics}}                                                                                                                                                                                                                                                                                                                                                                                      \\ \hline
\textbf{COT}    & 54.09                                                & 38.39                                                    & 19.59                                                & 28.81                                                & 22.98                                                     & 28.02                                                 & 41.67                                               \\
\textbf{EER}    & 42.50                                                & 23.66                                                    & 21.65                                                & 24.15                                                & 20.97                                                     & 24.74                                                 & 41.67                                               \\ \hline
\end{tabular}%
}
\caption{ Table showing the performance of different models across prompting strategies. The models were evaluated on annotated maps of China. Here "Acc" stands for accuracy}
\vspace{-1.0em}
\label{tab:prompt}
\end{table}

\paragraph{Prompt Effectiveness.}
While most models consistently perform better with the standard Chain-of-Thought (COT) prompt compared to the Explicit Extraction and Reasoning (EER) prompt (as evident in Table \ref{tab:prompt}), a notable exception is Gemini 1.5 Flash, which performs comparably or even better with the EER prompt.  This suggests that Gemini possesses particularly strong instruction-following capabilities.  Smaller, open-source models likely struggle with following the complex, step-wise instructions within the EER prompt.  However, analysis of responses from larger models reveals that they implicitly adopt a methodology similar to EER, demonstrating impressive progress in their reasoning abilities and mimicking human-like thinking.

\section{Biases in Model Prediction}
This section analyzes the performance variations of models across different map and question variants.  While these observations are often influenced by question type, we highlight the most prominent insights.

\subsection{Map Variants}
\paragraph{Discrete vs. Continuous Maps.}
While it is challenging to directly compare model performance on continuous and discrete maps due to the differing question types, a general trend emerges: models tend to perform better on discrete maps (as shown in Table \ref{tab:dch}). This trend is particularly pronounced for questions involving counting and extracting ranges, suggesting that models might struggle with accurately extracting legend ranges and color resolution in continuous maps.  Interestingly, models performed significantly better on single-word answers within the continuous category. This may be attributed to the simplicity of these questions, as the task itself is inherently challenging for humans.

\begin{table}[ht!]
\resizebox{\columnwidth}{!}{%
\begin{tabular}{cccccccc}
\hline
\textbf{\begin{tabular}[c]{@{}c@{}}Map \\ Type\end{tabular}}          & \textbf{\begin{tabular}[c]{@{}c@{}}Binary \\ Acc\end{tabular}} & \textbf{\begin{tabular}[c]{@{}c@{}}Single \\ Recall \end{tabular}} & \textbf{\begin{tabular}[c]{@{}c@{}}Count \\ Acc\end{tabular}} & \textbf{\begin{tabular}[c]{@{}c@{}}Range \\ Acc\end{tabular}} & \textbf{\begin{tabular}[c]{@{}c@{}}List \\ Precision\end{tabular}} & \textbf{\begin{tabular}[c]{@{}c@{}}List \\ Recall\end{tabular}} & \textbf{\begin{tabular}[c]{@{}c@{}}Rank\\ RWP\end{tabular}} \\ \hline
\multicolumn{8}{c}{\textbf{GPT-4o}}                                                                             \\ \hline
\textbf{with}    & 71.52     & 40.06       & 35.48    & 55.75    & 49.94    & 49.17      & 53.70  \\
\textbf{without} & 66.45     & 40.92       & 30.85    & 53.54    & 46.23    & 47.09      & 55.56  \\ \hline
\multicolumn{8}{c}{\textbf{Gemini 1.5 Flash}}                                                                   \\ \hline
\textbf{with}    & 56.36     & 38.49       & 24.47    & 40.27    & 34.55    & 45.11      & 38.69  \\
\textbf{without} & 58.99     & 37.39       & 23.40    & 35.84    & 36.25    & 46.21      & 45.83  \\ \hline
\multicolumn{8}{c}{\textbf{InternLM-XComposer2}}                                                                \\ \hline
\textbf{with}    & 56.80     & 32.37       & 17.02    & 13.27    & 20.14    & 24.02      & 35.71  \\
\textbf{without} & 53.51     & 34.08       & 14.89    & 13.27    & 27.84    & 35.84      & 39.29  \\ \hline
\multicolumn{8}{c}{\textbf{Idefics}}                                                                            \\ \hline
\textbf{with}    & 54.17     & 43.59       & 13.83    & 28.76    & 32.19    & 38.29      & 45.24  \\
\textbf{without} & 56.58     & 43.48       & 15.96    & 23.45    & 28.04    & 36.26      & 48.81  \\ \hline
\end{tabular}%
}
\caption{Table showing the performance of different models across maps of India with and without annotations, using the zero-shot COT prompt. Here, "Acc" stands for accuracy, "with" and "without" represent the presence and absence of annotations respectively.}
\vspace{-1.0em}
\label{tab:annot}
\end{table}

\paragraph{Maps with and without annotations.}
As shown in Table \ref{tab:annot}, models generally exhibited similar performance on maps with and without annotations, with only a slight improvement observed for annotated maps in some cases.  Surprisingly, we also found instances where models performed better on maps without annotations.  This suggests that while annotations can be beneficial, they are not a critical factor in building models for understanding maps.
\vspace{-0.1cm}

\begin{table}[ht!]
\resizebox{\columnwidth}{!}{%
\begin{tabular}{ccccccc}
\hline
\textbf{\begin{tabular}[c]{@{}c@{}}Map \\ Type\end{tabular}}          & \textbf{\begin{tabular}[c]{@{}c@{}}Binary \\ Acc\end{tabular}} & \textbf{\begin{tabular}[c]{@{}c@{}}Single \\ Recall \end{tabular}} & \textbf{\begin{tabular}[c]{@{}c@{}}Count \\ Acc\end{tabular}} & \textbf{\begin{tabular}[c]{@{}c@{}}Range \\ Acc\end{tabular}} & \textbf{\begin{tabular}[c]{@{}c@{}}List \\ Precision\end{tabular}} & \textbf{\begin{tabular}[c]{@{}c@{}}List \\ Recall\end{tabular}} \\ \hline
\multicolumn{7}{c}{\textbf{GPT-4o}}                                                                     \\ \hline
\textbf{continuous} & 64.05     & 62.12       & 25.00    & 61.84    & 36.70    & 42.11      \\
\textbf{discrete}   & 73.72     & 35.62       & 56.10    & 74.76    & 39.85    & 47.80      \\
\textbf{hatched}    & 49.78     & 16.14       & 26.83    & 26.67    & 26.96    & 32.60      \\ \hline
\multicolumn{7}{c}{\textbf{Gemini 1.5 Flash}}                                                           \\ \hline
\textbf{continuous} & 63.03     & 56.52       & 25.00    & 43.75    & 38.84    & 53.39      \\
\textbf{discrete}   & 66.20     & 53.64       & 56.10    & 70.48    & 38.66    & 50.26      \\
\textbf{hatched}    & 49.36     & 56.20       & 51.22    & 53.95    & 20.51    & 35.35      \\ \hline
\multicolumn{7}{c}{\textbf{InternLM-XComposer2}}                                                        \\ \hline
\textbf{continuous} & 54.55     & 50.72       & 12.50    & 22.50    & 23.56    & 34.18      \\
\textbf{discrete}   & 51.76     & 27.91       & 36.59    & 19.05    & 22.08    & 32.41      \\
\textbf{hatched}    & 45.49     & 31.40       & 53.66    & 17.11    & 25.84    & 31.31      \\ \hline
\multicolumn{7}{c}{\textbf{Idefics}}                                                                    \\ \hline
\textbf{continuous} & 61.21     & 63.77       & 12.50    & 27.50    & 23.55    & 41.24      \\
\textbf{discrete}   & 51.06     & 59.69       & 19.51    & 30.48    & 28.48    & 44.02      \\
\textbf{hatched}    & 52.79     & 56.98       & 12.20    & 25.00    & 22.99    & 42.09      \\ \hline
\end{tabular}%
}
\caption{Table showing the performance of different models across discrete, continuous and hatched maps of USA, using the zero-shot COT prompt. Here "Acc" stands for Accuracy.}
\vspace{-1.0em}
\label{tab:dch}
\end{table}

\paragraph{Colored Maps vs. Hatched Maps.}
All models consistently performed better on colored maps compared to hatched maps, demonstrating a preference for colored depictions of data (as seen in Table \ref{tab:dch}). This trend is notable, as even models like GPT-4o experienced significant score drops on hatched maps, highlighting a lack of robustness.  Impressively, Idefics displayed the least performance decline, suggesting a more robust ability to accurately extract data from these visually complex maps.

\subsection{Country-Wise Performance}
Table \ref{tab:country-wise} presents model performance across different countries.  While a consistent pattern is difficult to discern, a notable trend emerges: open-source models generally demonstrate consistent performance across countries, while closed-source models exhibit greater variation.  The exact cause of this variation remains unclear, but potential contributing factors include biases in the training data.

\begin{table}[ht!]
\resizebox{\columnwidth}{!}{%
\begin{tabular}{cccccccc}
\hline
\textbf{\begin{tabular}[c]{@{}c@{}}Map \\ Type\end{tabular}}          & \textbf{\begin{tabular}[c]{@{}c@{}}Binary \\ Acc\end{tabular}} & \textbf{\begin{tabular}[c]{@{}c@{}}Single \\ Recall \end{tabular}} & \textbf{\begin{tabular}[c]{@{}c@{}}Count \\ Acc\end{tabular}} & \textbf{\begin{tabular}[c]{@{}c@{}}Range \\ Acc\end{tabular}} & \textbf{\begin{tabular}[c]{@{}c@{}}List \\ Precision\end{tabular}} & \textbf{\begin{tabular}[c]{@{}c@{}}List \\ Recall\end{tabular}} & \textbf{\begin{tabular}[c]{@{}c@{}}Rank\\ RWP\end{tabular}} \\ \hline
\multicolumn{8}{c}{\textbf{GPT-4o}}                                                                             \\ \hline
\textbf{USA}     & 70.26       & 44.68         & 51.02      & 71.28      & 38.64      & 45.61        & -        \\
\textbf{India}   & 71.52       & 40.06         & 35.48      & 55.75      & 49.94      & 49.17        & 54.94    \\
\textbf{China}   & 66.97       & 47.53         & 50.52      & 59.40      & 53.93      & 57.56        & 45.66    \\ \hline
\multicolumn{8}{c}{\textbf{Gemini 1.5 Flash}}                                                                   \\ \hline
\textbf{USA}     & 65.03       & 54.65         & 51.02      & 63.10      & 38.73      & 51.43        & -        \\
\textbf{India}   & 56.36       & 38.49         & 24.47      & 40.27      & 34.55      & 45.11        & 38.99    \\
\textbf{China}   & 62.27       & 51.83         & 13.40      & 52.97      & 22.76      & 38.96        & 54.31    \\ \hline
\multicolumn{8}{c}{\textbf{InternLM-XComposer2}}                                                                \\ \hline
\textbf{USA}     & 52.78       & 35.86         & 32.65      & 20.00      & 22.63      & 33.07        & -        \\
\textbf{India}   & 56.80       & 32.37         & 17.02      & 13.27      & 20.14      & 24.02        & 35.71    \\
\textbf{China}   & 54.09       & 38.39         & 19.59      & 28.81      & 22.98      & 28.02        & 41.99    \\ \hline
\multicolumn{8}{c}{\textbf{Idefics}}                                                                            \\ \hline
\textbf{USA}     & 54.79       & 61.11         & 18.37      & 29.66      & 26.64      & 42.99        & -        \\
\textbf{India}   & 54.17       & 43.59         & 13.83      & 28.76      & 32.19      & 38.29        & 45.24    \\
\textbf{China}   & 54.09       & 50.54         & 21.65      & 34.32      & 21.67      & 29.91        & 46.15    \\ \hline
\end{tabular}%
}
\caption{Table showing model performance across annotated maps of USA, India, China, using the zero-shot COT prompt. Here "Acc" denotes accuracy. }
\vspace{-1.0em}
\label{tab:country-wise}
\end{table}

\subsection{Analysis across Question and Answer Types}
Table \ref{tab:country-wise} reveals that models generally performed best on questions requiring a binary answer, followed by single-word answers, highlighting their strong data extraction capabilities.  Closed-source models like Gemini and GPT also excelled at questions expecting a range; however, smaller models struggled in this domain, likely due to limited reasoning or color extraction skills.  Models encountered the most difficulty with tasks requiring a count or listing, which demand complex reasoning, external knowledge, and geospatial understanding.  These questions proved challenging not only for models but also for humans (as shown in Table \ref{tab:human_baseline_results}).  For questions concerning relative regions, models struggled with single-word or count-based answers, further highlighting the complexity of these tasks, which require external knowledge, relative region extraction, and complex reasoning.  Smaller models, in particular, struggled in this category (as seen in the Appendix \ref{sec:comprehensive_results} tables for relative regions).

\begin{figure*}[ht]
  \centering
\includegraphics[width=6in]{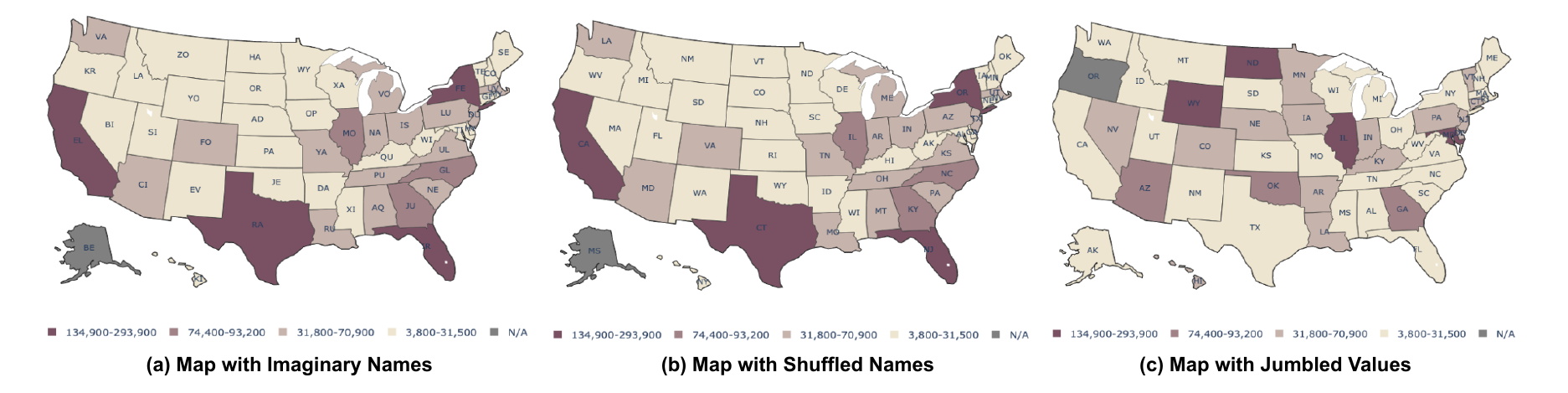}
\captionsetup{justification=centering}
  \caption{Examples of map with Imaginary and Shuffled names and Jumbled Values for the same underlying data.}
  \label{fig:counter_factual_example}
\vspace{-1.0em}
\end{figure*}

\section{Human Evaluation and Baseline}

We conducted a human evaluation of the MapWise dataset to establish a human baseline and to compare the performance of models against human evaluators. The MapWise dataset is particularly challenging as it demands careful identification of subtle shades and patterns, as well as nuanced understanding of spatial geographical relationships. We conducted the human evaluation on a uniformly sampled set of 150 unique questions, spanning 75 maps and 40 templates. We ensured an approximately equal distribution of each answer type and map type, further ensuring the proper representation of continuous and discrete maps and relative region-type questions. This approach was followed for all three countries to capture all diverse scenarios within the dataset. We employed majority voting for result verification of the three independent annotators.

\vspace{-0.5em}
\begin{table}[ht!]
\resizebox{\columnwidth}{!}{%
\begin{tabular}{cccccccc}
\hline

\multirow{2}{*}{\textbf{Country}} & \textbf{Binary} & \textbf{Single} & \textbf{Count} & \textbf{Range} & \textbf{List} & \textbf{List} &\textbf{Rank} \\
& \textbf{Acc.} & \textbf{Recall} & \textbf{Acc.} & \textbf{Acc} & \textbf{Prec.} & \textbf{Recall} & \textbf{RWP}\\ \hline

\toprule
USA &94.74 &96.67 &88.89 &100.00 &95.16 &93.55 &-/- \\
India &96.97 &86.21 &80.00 &89.29 &98.61 &94.44 &91.67 \\
China &100.00 &88.99 &79.31 &80.77 &79.76 &79.76 &80.00 \\
\bottomrule

\end{tabular}%
}
\caption{Human Baseline results (in \%), Acc stands for Accuracy.}
\vspace{-1.0em}
\label{tab:human_baseline_results} 
\end{table}

As shown in Table \ref{tab:human_baseline_results}, the less-than-perfect human performance highlights the complexity of the task and offers a realistic benchmark against which model performance can be compared. Several common challenges contribute to the dataset's complexity, even for human evaluators. These include confusing color shades, particularly in continuous maps, numerous range groups in discrete maps, difficulty in understanding patterns for hatched maps and the challenge of accurately interpreting values for regions with smaller areas.

\vspace{-0.5em}
\begin{table}[ht!]
\resizebox{\columnwidth}{!}{%
\begin{tabular}{cccccccc}
\hline

\multirow{2}{*}{\textbf{Country}} & \textbf{Binary} & \textbf{Single} & \textbf{Count} & \textbf{Range} & \textbf{List} & \textbf{Rank} \\
& \textbf{(yes/no)} & \textbf{} & \textbf{Integer} & \textbf{A-B, >A, <B} & \textbf{} & \textbf{} \\ \hline

\toprule
USA &100.00 &96.67 &88.89 &100.00 &96.77 &-/-  \\
India &96.97 &89.66 &86.67 &100.00 &100.00 &100.00  \\
China &100.00 &96.43 &89.66 &100.00 &100.00 &80.00  \\
\bottomrule

\end{tabular}%
}
\caption{Percentage of Responses which has Majority}
\vspace{-1.0em}
\label{tab:human_baseline_majority_perc} 
\end{table}

From Table \ref{tab:human_baseline_majority_perc}, we observe that for binary, range and list type answer, there is nearly 100\% majority agreement among human evaluators. However, there is a slight decline in majority agreement for single type answers and least majority for count type answer, highlighting the confusion and variability in responses among human evaluators. 

\section{Experiments with Counterfactual data}
We performed additional analysis to evaluate models which are trained extensively on large datasets, under conditions where their internal factual knowledge was limited. To carry out our analysis, we created three types of counterfactual data that forced the models to rely exclusively on the provided maps. Figure \ref{fig:counter_factual_example} shows examples of our counterfactual maps.

For the counterfactual dataset generation, we first uniformly sampled a subset of 240 unique Questions from USA dataset, spreading over 90 Maps and 26 Templates. We also ensured approximately equal distribution of each answer type. Using the sampled dataset as a representative sample (consisting of original names and values), we applied the following modifications to create our countefactual dataset: 

\vspace{-0.5em}
\paragraph{\textit{Imaginary names.}}
States were assigned imaginary names, generated using GPT-4.  
(\textit{e.g., Alabama was renamed Aquilis, Arkansas became Davina, etc.}) The first two letters of these imaginary names were used as state codes for an annotated map of the US.

\vspace{-0.5em}
\paragraph{\textit{Shuffled names.}}
The names of different US states were randomly shuffled while retaining the values of each geographical region. Annotated maps with these shuffled state codes were generated (e.g. Alabama became Montana, Arkansas became Idaho). 

\vspace{-0.5em}
\paragraph{\textit{Jumbled values.}}
The values corresponding to each of the different US states were shuffled, keeping the legend fixed. As a result, several question answer pairs needed to be re-evaluated.

\begin{table}[ht!]
\resizebox{\columnwidth}{!}{%
\begin{tabular}{ccccccc}
\hline
\multirow{2}{*}{\textbf{CF Type}} & \textbf{Binary} & \textbf{Single} & \textbf{Count} & \textbf{Range} & \textbf{List} & \textbf{List} \\
& \textbf{Acc.} & \textbf{Recall} & \textbf{Acc.} & \textbf{Acc} & \textbf{Prec.} & \textbf{Recall} \\ \hline

\multicolumn{7}{c}{\textbf{Gemini 1.5 Flash}} \\ \hline
Original &59.18 &35.42 &20.00 &35.42 &47.52 &60.20 \\
Imaginary &53.06 &23.96 &11.11 &35.42 &24.86 &38.27 \\
Shuffled &63.27 &25.00 &22.22 &37.50 &18.84 &25.68 \\
Jumbled &53.06 &30.21 &31.11 &40.63 &39.88 &45.41 \\ \hline
\multicolumn{7}{c}{\textbf{GPT 4o}} \\ \hline
Original &61.12 &37.48 &22.11 &36.99 &49.58 &62.25 \\
Imaginary &55.09 &25.98 &13.13 &37.46 &26.89 &40.29 \\
Shuffled &65.31 &27.03 &24.24 &39.53 &20.87 &28.72 \\
Jumbled &55.09 &32.24 &33.13 &42.67 &41.92 &47.45 \\ \hline
\multicolumn{7}{c}{\textbf{Idefics}} \\ \hline
Original &55.10 &31.25 &13.33 &30.21 &26.19 &47.79 \\
Imaginary &46.94 &0.00 &8.89 &16.67 &0.00 &0.00 \\
Shuffled &53.06 &12.50 &13.33 &25.00 &7.82 &13.95 \\
Jumbled &32.65 &14.58 &11.11 &23.96 &25.83 &43.88 \\ \hline
\multicolumn{7}{c}{\textbf{InternLM}} \\ \hline
Original &46.94 &13.54 &28.89 &14.58 &26.17 &40.82 \\
Imaginary &53.06 &0.00 &20.00 &8.33 &3.96 &9.86 \\
Shuffled &55.10 &10.42 &15.56 &13.54 &11.85 &15.31 \\
Jumbled &42.86 &13.54 &15.56 &6.25 &22.59 &30.78 \\
\bottomrule
\end{tabular}%
}
\caption{Counter Factual Results (in \%) for zero-shot COT prompt. CF represents Counter Factual and Acc. stands for Accuracy.}
\vspace{-1.0em}
\label{tab:counterFactual_mainresults} 
\end{table}

% \subsection{Results and Analysis}
Adjustments to the prompts were made in accordance with the specific requirements of each counterfactual dataset. For example when dealing with imaginary names, the following instruction was included: \textit{"The map in the image represents fictional names for each state as specified in the following dictionary. Use this dictionary while analyzing the map"}. A corresponding dictionary was provided for reference within the prompt. Table \ref{tab:counterFactual_mainresults} presents the results for Gemini, GPT, Idefics and InternLM, evaluated using the zero-shot COT prompt (Appendix \ref{sec:rem_jumbled_app} for contains results for the remaining models and the EER prompt). At a high level, it is evident that the closed source model consistently outperformed the open-source models across all three types of counterfactual datasets.

Upon closer inspection, we notice a significant decline in performance for Single and List type answers when using imaginary and shuffled names compared to the original dataset. However, the comparable or better results for Binary, Count and Range type suggest that models are usually able to follow instruction, but tend to diverge while generating the counterfactual names, often relying on internal knowledge or producing hallucinated responses, despite explicit instruction to avoid this behavior. In the case of imaginary names, the open source models attain scores close to 0, indicating their inability to generate counterfactual names. Upon reviewing the responses, it was evident that while these models initiate a reasoning, they almost always hallucinate when generating the counterfactual state names. Notably, we also see a drop in questions with jumbled values, emphasizing the correlation between values and their corresponding states.
\looseness -1

\section{Related Work}

\textbf{Visual Question Answering (VQA)} has attracted significant attention in computer vision and natural language processing due to its interdisciplinary challenges, as explored by \citet{antol2015vqa, goyal2017making, bazi2023vision, hartsock2024vision, zhang2024vision}. The introduction of Visual Question Rewriting (VQR) by \citet{10.1145/3404835.3463114} has further advanced our understanding of how visual information can enhance question-answering systems. Similarly, \citet{10.2478/amns.2023.1.00182} introduced visual quizzing, which involves reasoning with both images and their related questions.

\textbf{Map Question Answering (MQA)} and \textbf{Chart Question Answering (CQA)} have also emerged as challenging extensions of VQA, requiring the interpretation of visual data representations such as charts and maps. Datasets like ChartQA\citep{kafle2018dvqa, kahou2017figureqa} focus on interpreting structured data charts, while \citet{chang2022mapqa} introduced MapQA for choropleth map question answering, highlighting the need for robust VQA systems. MapQA's U.S. focus study and template questions limit its scope. Our dataset on the other hand includes a diverse set of countries, map types and complex questions which were manually curated to create an effective benchmark to evaluate model performances. 

\textbf{Enhancing Visual Question Answering.} Despite these advances, gaps remain in Chart (CQA) and Map Question Answering (MQA), particularly in handling complex reasoning, numeric answers, and out-of-vocabulary terms. Existing systems often struggle with these challenges, and synthetic datasets may limit their real-world applicability \cite{10.18653/v1/2023.findings-eacl.189, chaudhry2020leaf}. Our research addresses these issues by building on \citet{chang2022mapqa} with more diverse maps, challenging questions, and benchmarking state-of-the-art multimodal and visual-language models.

\section{Conclusion and Future Work}
% This paper introduces \textbf{MAPWise}, a new large-scale dataset for understanding choropleth maps in the United States, China, and India. Future research could broaden the scope of datasets by including different map types, such as fictional maps or detailed maps with features like rivers and roads, to evaluate VLM generalization across geographical contexts.  Further research is needed to identify and mitigate biases inherent in map interpretation. Techniques like dataset perturbation could provide deeper insights and help mitigate biases effectively. 

% To improve data extraction, integrating external knowledge sources, like RAG networks filled with detailed information about state borders and regional relationships, could enhance VLM reasoning.  Another future direction would be improving VLM color recognition accuracy and integrating additional datasets, such as charts, to enhance their ability to interpret and process map-related information effectively.
This paper introduces \textbf{MAPWise}, a new large-scale dataset tailored for understanding choropleth maps in three diverse countries: the United States, China, and India. Looking ahead, there are many promising areas for further research based on what we found and from the existing studies. Future studies could broaden the scope of datasets by including different types of maps. Inspired by previous work \cite{fan2024understanding}, we could complement our dataset by exploring fictional maps or more detailed maps that include features such as rivers and roads. This expansion would help evaluate how well VLMs generalize across diverse geographical contexts. Further research is needed to identify and mitigate biases inherent in map interpretation. Techniques like dataset perturbation, which introduces variations in map features and contexts, could provide deeper insights and help mitigate biases effectively.

To improve how data is extracted, integrating external knowledge sources in future would be a promising strategy. Models that use knowledge graphs, like RAG networks filled with detailed information about state borders and regional relationships, could also improve how well Vision Language Models (VLMs) reason through map-based tasks. Another future direction would be improving how VLMs are trained to recognize colors more accurately and integrating additional datasets, training on auxiliary data such as charts, to improve their ability to interpret and process map-related information effectively.

\section*{Limitations}
While our study has yielded interesting observations, it's crucial to acknowledge its limitations. We focused exclusively on choropleth maps, which represent data using color gradients. While these maps are effective for visualizing regional data, they lack the detailed features and interactive elements found in more advanced mapping systems like Google Maps. 

Additionally, our study does not include rank-based questions specifically tailored for the United States. Therefore, our findings and methods may not fully generalize to these more complex mapping systems and their unique challenges. Moreover, we were limited to maps from only three countries, and the manual question creation process restricted the size of our dataset.

\section*{Ethics Statement}
We, the authors, ensure that our research meets the highest ethical standards in both research and publication. We have carefully addressed all ethical considerations for responsible and fair use of computational linguistics methods. To help others replicate our results, we are sharing all necessary details, including code, available datasets (used according to their ethical guidelines), and other resources. This allows the research community to verify and build on our work. Our claims are backed by our experimental results. We provide detailed information on annotations, dataset splits, models, and methods used for reproducibility.

\section*{Acknowledgement}
Our work is sponsored by the Army Research Office and is accomplished under Grant Number W911NF-20-1-0080. The views and conclusions contained in this document are those of the authors and should not be interpreted as representing the official policies, either expressed or implied, of the Army Research Office or the U.S. Government. The U.S. Government is authorized to reproduce and distribute reprints for Government purposes notwithstanding any copyright notation herein. This work was partially funded by ONR Contract N00014-19-1-2620. We would also like to thank Nirupama Ratna, Arqam Patel and Jay Gala for their extensive help and support during the process of creating our dataset. Additionally, we would like to extend our deepest gratitude to Manish Shrivastava for his guidance during the project. 

\bibliography{custom}

\begin{thebibliography}{24}
\providecommand{\natexlab}[1]{#1}

\bibitem[{Antol et~al.(2015)Antol, Agrawal, Lu, Mitchell, Batra, Zitnick, and Parikh}]{antol2015vqa}
Stanislaw Antol, Aishwarya Agrawal, Jiasen Lu, Margaret Mitchell, Dhruv Batra, C~Lawrence Zitnick, and Devi Parikh. 2015.
\newblock Vqa: Visual question answering.
\newblock In \emph{Proceedings of the IEEE international conference on computer vision}, pages 2425--2433.

\bibitem[{Bai et~al.(2023{\natexlab{a}})Bai, Bai, Chu, Cui, Dang, Deng, Fan, Ge, Han, Huang, Hui, Ji, Li, Lin, Lin, Liu, Liu, Lu, Lu, Ma, Men, Ren, Ren, Tan, Tan, Tu, Wang, Wang, Wang, Wu, Xu, Xu, Yang, Yang, Yang, Yang, Yao, Yu, Yuan, Yuan, Zhang, Zhang, Zhang, Zhang, Zhou, Zhou, Zhou, and Zhu}]{bai2023qwen}
Jinze Bai, Shuai Bai, Yunfei Chu, Zeyu Cui, Kai Dang, Xiaodong Deng, Yang Fan, Wenbin Ge, Yu~Han, Fei Huang, Binyuan Hui, Luo Ji, Mei Li, Junyang Lin, Runji Lin, Dayiheng Liu, Gao Liu, Chengqiang Lu, Keming Lu, Jianxin Ma, Rui Men, Xingzhang Ren, Xuancheng Ren, Chuanqi Tan, Sinan Tan, Jianhong Tu, Peng Wang, Shijie Wang, Wei Wang, Shengguang Wu, Benfeng Xu, Jin Xu, An~Yang, Hao Yang, Jian Yang, Shusheng Yang, Yang Yao, Bowen Yu, Hongyi Yuan, Zheng Yuan, Jianwei Zhang, Xingxuan Zhang, Yichang Zhang, Zhenru Zhang, Chang Zhou, Jingren Zhou, Xiaohuan Zhou, and Tianhang Zhu. 2023{\natexlab{a}}.
\newblock \href {https://arxiv.org/abs/2309.16609} {Qwen technical report}.
\newblock \emph{Preprint}, arXiv:2309.16609.

\bibitem[{Bai et~al.(2023{\natexlab{b}})Bai, Bai, Yang, Wang, Tan, Wang, Lin, Zhou, and Zhou}]{bai2023qwenvl}
Jinze Bai, Shuai Bai, Shusheng Yang, Shijie Wang, Sinan Tan, Peng Wang, Junyang Lin, Chang Zhou, and Jingren Zhou. 2023{\natexlab{b}}.
\newblock \href {https://arxiv.org/abs/2308.12966} {Qwen-vl: A versatile vision-language model for understanding, localization, text reading, and beyond}.
\newblock \emph{Preprint}, arXiv:2308.12966.

\bibitem[{Bazi et~al.(2023)Bazi, Rahhal, Bashmal, and Zuair}]{bazi2023vision}
Yakoub Bazi, Mohamad Mahmoud~Al Rahhal, Laila Bashmal, and Mansour Zuair. 2023.
\newblock Vision--language model for visual question answering in medical imagery.
\newblock \emph{Bioengineering}, 10(3):380.

\bibitem[{Bhaisaheb et~al.(2023)Bhaisaheb, Paliwal, Patil, Patwardhan, Vig, and Shroff}]{10.18653/v1/2023.findings-eacl.189}
S.~Bhaisaheb, S.~Paliwal, R.~Patil, M.~Patwardhan, L.~Vig, and G.~Shroff. 2023.
\newblock \href {https://doi.org/10.18653/v1/2023.findings-eacl.189} {Program synthesis for complex qa on charts via probabilistic grammar based filtered iterative back-translation}.
\newblock \emph{Findings of the Association for Computational Linguistics: EACL 2023}.

\bibitem[{Cai et~al.(2024)Cai, Cao, Chen, Chen, Chen, Chen, Chen, Chen, Chen, Chu, Dong, Duan, Fan, Fei, Gao, Ge, Gu, Gu, Gui, Guo, Guo, He, Hu, Huang, Jiang, Jiao, Jin, Lei, Li, Li, Li, Li, Li, Li, Liu, Liu, Hong, Liu, Liu, Liu, Lv, Lv, Lv, Ma, Ma, Ma, Ning, Ouyang, Qiu, Qu, Shang, Shao, Song, Song, Sui, Sun, Sun, Tang, Wang, Wang, Wang, Wang, Wang, Wang, Wang, Wei, Weng, Wu, Xiong, Xu, Xu, Yan, Yan, Yang, Ye, Ying, Yu, Yu, Zang, Zhang, Zhang, Zhang, Zhang, Zhang, Zhang, Zhang, Zhang, Zhang, Zhang, Zhang, Zhao, Zhao, Zhao, Zhou, Zhou, Zhuo, Zou, Qiu, Qiao, and Lin}]{cai2024internlm2}
Zheng Cai, Maosong Cao, Haojiong Chen, Kai Chen, Keyu Chen, Xin Chen, Xun Chen, Zehui Chen, Zhi Chen, Pei Chu, Xiaoyi Dong, Haodong Duan, Qi~Fan, Zhaoye Fei, Yang Gao, Jiaye Ge, Chenya Gu, Yuzhe Gu, Tao Gui, Aijia Guo, Qipeng Guo, Conghui He, Yingfan Hu, Ting Huang, Tao Jiang, Penglong Jiao, Zhenjiang Jin, Zhikai Lei, Jiaxing Li, Jingwen Li, Linyang Li, Shuaibin Li, Wei Li, Yining Li, Hongwei Liu, Jiangning Liu, Jiawei Hong, Kaiwen Liu, Kuikun Liu, Xiaoran Liu, Chengqi Lv, Haijun Lv, Kai Lv, Li~Ma, Runyuan Ma, Zerun Ma, Wenchang Ning, Linke Ouyang, Jiantao Qiu, Yuan Qu, Fukai Shang, Yunfan Shao, Demin Song, Zifan Song, Zhihao Sui, Peng Sun, Yu~Sun, Huanze Tang, Bin Wang, Guoteng Wang, Jiaqi Wang, Jiayu Wang, Rui Wang, Yudong Wang, Ziyi Wang, Xingjian Wei, Qizhen Weng, Fan Wu, Yingtong Xiong, Chao Xu, Ruiliang Xu, Hang Yan, Yirong Yan, Xiaogui Yang, Haochen Ye, Huaiyuan Ying, Jia Yu, Jing Yu, Yuhang Zang, Chuyu Zhang, Li~Zhang, Pan Zhang, Peng Zhang, Ruijie Zhang, Shuo Zhang, Songyang Zhang, Wenjian Zhang,
  Wenwei Zhang, Xingcheng Zhang, Xinyue Zhang, Hui Zhao, Qian Zhao, Xiaomeng Zhao, Fengzhe Zhou, Zaida Zhou, Jingming Zhuo, Yicheng Zou, Xipeng Qiu, Yu~Qiao, and Dahua Lin. 2024.
\newblock \href {https://arxiv.org/abs/2403.17297} {Internlm2 technical report}.
\newblock \emph{Preprint}, arXiv:2403.17297.

\bibitem[{Chang et~al.(2022)Chang, Palzer, Li, Fosler-Lussier, and Xiao}]{chang2022mapqa}
Shuaichen Chang, David Palzer, Jialin Li, Eric Fosler-Lussier, and Ningchuan Xiao. 2022.
\newblock Mapqa: A dataset for question answering on choropleth maps.
\newblock \emph{arXiv preprint arXiv:2211.08545}.

\bibitem[{Chaudhry et~al.(2020)Chaudhry, Shekhar, Gupta, Maneriker, Bansal, and Joshi}]{chaudhry2020leaf}
Ritwick Chaudhry, Sumit Shekhar, Utkarsh Gupta, Pranav Maneriker, Prann Bansal, and Ajay Joshi. 2020.
\newblock Leaf-qa: Locate, encode \& attend for figure question answering.
\newblock In \emph{Proceedings of the IEEE/CVF Winter Conference on Applications of Computer Vision}, pages 3512--3521.

\bibitem[{Dong et~al.(2024)Dong, Zhang, Zang, Cao, Wang, Ouyang, Wei, Zhang, Duan, Cao, Zhang, Li, Yan, Gao, Zhang, Li, Li, Chen, He, Zhang, Qiao, Lin, and Wang}]{dong2024internlmxcomposer2}
Xiaoyi Dong, Pan Zhang, Yuhang Zang, Yuhang Cao, Bin Wang, Linke Ouyang, Xilin Wei, Songyang Zhang, Haodong Duan, Maosong Cao, Wenwei Zhang, Yining Li, Hang Yan, Yang Gao, Xinyue Zhang, Wei Li, Jingwen Li, Kai Chen, Conghui He, Xingcheng Zhang, Yu~Qiao, Dahua Lin, and Jiaqi Wang. 2024.
\newblock \href {https://arxiv.org/abs/2401.16420} {Internlm-xcomposer2: Mastering free-form text-image composition and comprehension in vision-language large model}.
\newblock \emph{Preprint}, arXiv:2401.16420.

\bibitem[{Fan et~al.(2024)Fan, Lei, Mancenido, Maceachren, and Maciejewski}]{fan2024understanding}
Arlen Fan, Fan Lei, Michelle Mancenido, Alan~M Maceachren, and Ross Maciejewski. 2024.
\newblock Understanding reader takeaways in thematic maps under varying text, detail, and spatial autocorrelation.
\newblock In \emph{Proceedings of the CHI Conference on Human Factors in Computing Systems}, pages 1--17.

\bibitem[{Gemini(2024)}]{geminiteam2024gemini}
Team Gemini. 2024.
\newblock \href {https://arxiv.org/abs/2403.05530} {Gemini 1.5: Unlocking multimodal understanding across millions of tokens of context}.
\newblock \emph{Preprint}, arXiv:2403.05530.

\bibitem[{Goyal et~al.(2017)Goyal, Khot, Summers-Stay, Batra, and Parikh}]{goyal2017making}
Yash Goyal, Tejas Khot, Douglas Summers-Stay, Dhruv Batra, and Devi Parikh. 2017.
\newblock Making the v in vqa matter: Elevating the role of image understanding in visual question answering.
\newblock In \emph{Proceedings of the IEEE conference on computer vision and pattern recognition}, pages 6904--6913.

\bibitem[{Hartsock and Rasool(2024)}]{hartsock2024vision}
Iryna Hartsock and Ghulam Rasool. 2024.
\newblock Vision-language models for medical report generation and visual question answering: A review.
\newblock \emph{arXiv preprint arXiv:2403.02469}.

\bibitem[{Hong et~al.(2023)Hong, Wang, Lv, Xu, Yu, Ji, Wang, Wang, Zhang, Li, Xu, Dong, Ding, and Tang}]{hong2023cogagent}
Wenyi Hong, Weihan Wang, Qingsong Lv, Jiazheng Xu, Wenmeng Yu, Junhui Ji, Yan Wang, Zihan Wang, Yuxuan Zhang, Juanzi Li, Bin Xu, Yuxiao Dong, Ming Ding, and Jie Tang. 2023.
\newblock \href {https://arxiv.org/abs/2312.08914} {Cogagent: A visual language model for gui agents}.
\newblock \emph{Preprint}, arXiv:2312.08914.

\bibitem[{Kafle et~al.(2018)Kafle, Price, Cohen, and Kanan}]{kafle2018dvqa}
Kushal Kafle, Brian Price, Scott Cohen, and Christopher Kanan. 2018.
\newblock Dvqa: Understanding data visualizations via question answering.
\newblock In \emph{Proceedings of the IEEE conference on computer vision and pattern recognition}, pages 5648--5656.

\bibitem[{Kahou et~al.(2017)Kahou, Michalski, Atkinson, K{\'a}d{\'a}r, Trischler, and Bengio}]{kahou2017figureqa}
Samira~Ebrahimi Kahou, Vincent Michalski, Adam Atkinson, {\'A}kos K{\'a}d{\'a}r, Adam Trischler, and Yoshua Bengio. 2017.
\newblock Figureqa: An annotated figure dataset for visual reasoning.
\newblock \emph{arXiv preprint arXiv:1710.07300}.

\bibitem[{OpenAI(2024)}]{openai2024gpt4}
OpenAI. 2024.
\newblock \href {https://arxiv.org/abs/2303.08774} {Gpt-4 technical report}.
\newblock \emph{Preprint}, arXiv:2303.08774.

\bibitem[{Salaberria et~al.(2023)Salaberria, Azkune, de~Lacalle, Soroa, and Agirre}]{salaberria2023image}
Ander Salaberria, Gorka Azkune, Oier~Lopez de~Lacalle, Aitor Soroa, and Eneko Agirre. 2023.
\newblock Image captioning for effective use of language models in knowledge-based visual question answering.
\newblock \emph{Expert Systems with Applications}, 212:118669.

\bibitem[{Wei et~al.(2021)Wei, Li, Zhang, and Wang}]{10.1145/3404835.3463114}
J.~Wei, X.~Li, Y.~Zhang, and X.~Wang. 2021.
\newblock \href {https://doi.org/10.1145/3404835.3463114} {Visual question rewriting for increasing response rate}.
\newblock \emph{Proceedings of the 44th International ACM SIGIR Conference on Research and Development in Information Retrieval}.

\bibitem[{Wei et~al.(2023)Wei, Wang, Schuurmans, Bosma, Ichter, Xia, Chi, Le, and Zhou}]{wei2023chainofthought}
Jason Wei, Xuezhi Wang, Dale Schuurmans, Maarten Bosma, Brian Ichter, Fei Xia, Ed~Chi, Quoc Le, and Denny Zhou. 2023.
\newblock \href {https://arxiv.org/abs/2201.11903} {Chain-of-thought prompting elicits reasoning in large language models}.
\newblock \emph{Preprint}, arXiv:2201.11903.

\bibitem[{Wu(2023)}]{10.2478/amns.2023.1.00182}
K.~Wu. 2023.
\newblock \href {https://doi.org/10.2478/amns.2023.1.00182} {Research and implementation of visual question and answer system based on deep learning}.
\newblock \emph{Applied Mathematics and Nonlinear Sciences}, 9.

\bibitem[{Yi et~al.(2024)Yi, Fan, Zhu, Lv, Liang, Wen, Pei, Li, and Tao}]{yi2024vlp2msa}
Guofeng Yi, Cunhang Fan, Kang Zhu, Zhao Lv, Shan Liang, Zhengqi Wen, Guanxiong Pei, Taihao Li, and Jianhua Tao. 2024.
\newblock Vlp2msa: expanding vision-language pre-training to multimodal sentiment analysis.
\newblock \emph{Knowledge-Based Systems}, 283:111136.

\bibitem[{Zhang et~al.(2024)Zhang, Huang, Jin, and Lu}]{zhang2024vision}
Jingyi Zhang, Jiaxing Huang, Sheng Jin, and Shijian Lu. 2024.
\newblock Vision-language models for vision tasks: A survey.
\newblock \emph{IEEE Transactions on Pattern Analysis and Machine Intelligence}.

\bibitem[{Zhao et~al.(2024)Zhao, Hao, Zi, Xu, and Wong}]{zhao2024bridging}
Shihao Zhao, Shaozhe Hao, Bojia Zi, Huaizhe Xu, and Kwan-Yee~K Wong. 2024.
\newblock Bridging different language models and generative vision models for text-to-image generation.
\newblock \emph{arXiv preprint arXiv:2403.07860}.

\end{thebibliography}
\newpage

\section*{Appendix}
\label{appendix}
\appendix

\section{Remaining Counter Factual Results}
\label{sec:rem_jumbled_app}
In this section we display the result for the remaining open-source models for zero-shot COT prompt (Table \ref{tab:counterFactual_cot_rem_models}) and results for all models for the EER prompt (Table \ref{tab: counterFactual_eer_allresults}) from the study. 

\begin{table}[ht!]
\resizebox{\columnwidth}{!}{%
\begin{tabular}{ccccccc}
\hline

\multirow{2}{*}{\textbf{CF Type}} & \textbf{Binary} & \textbf{Single} & \textbf{Count} & \textbf{Range} & \textbf{List} & \textbf{List} \\
& \textbf{Acc.} & \textbf{Recall} & \textbf{Acc.} & \textbf{Acc} & \textbf{Prec.} & \textbf{Recall} \\ \hline

\multicolumn{7}{c}{\textbf{CogAgent}} \\ \hline
Original &36.73  &19.79 &4.44 &10.42 &20.39 &40.14 \\
Imaginary &55.10  &0.00 &17.78 &10.42 &3.06 &3.06 \\
Shuffled &57.14  &17.71 &8.89 &5.21 &7.73 &9.52 \\
Jumbled &34.69  &15.63 &4.44 &6.25 &30.35 &43.54 \\ \hline
\multicolumn{7}{c}{\textbf{QwenVL}} \\ \hline
Original &48.98  &8.33 &15.56 &5.21 &15.84 &30.78 \\
Imaginary &47.92  &2.08 &13.33 &9.57 &3.01 &10.88 \\
Shuffled &51.02  &6.25 &11.11 &13.54 &6.95 &12.24 \\
Jumbled &38.78  &10.42 &4.44 &10.42 &25.24 &38.78 \\
\bottomrule

\end{tabular}%
}
\caption{Counter Factual Results (in \%) for zero-shot
COT prompt for CogAgent and QwenVL. CF represents Counter Factual and Acc.
stands for Accuracy.}
\label{tab:counterFactual_cot_rem_models} 
\end{table}

\begin{table}[ht!]
\resizebox{\columnwidth}{!}{%
\begin{tabular}{ccccccc}
\hline

\multirow{2}{*}{\textbf{CF Type}} & \textbf{Binary} & \textbf{Single} & \textbf{Count} & \textbf{Range} & \textbf{List} & \textbf{List} \\
& \textbf{Acc.} & \textbf{Recall} & \textbf{Acc.} & \textbf{Acc} & \textbf{Prec.} & \textbf{Recall} \\ \hline

\multicolumn{7}{c}{\textbf{Gemini 1.5 Flash}} \\ \hline
Original &68.75  &39.58 &26.67 &50.00 &43.25 &63.27 \\
Imaginary &61.22  &43.75 &20.00 &41.67 &24.91 &36.73 \\
Shuffled &71.43  &25.00 &13.33 &37.76 &6.80 &9.52 \\
Jumbled &66.67  &28.13 &13.33 &48.96 &43.58 &41.50 \\ \hline
\multicolumn{7}{c}{\textbf{GPT 4o}} \\ \hline
Original &70.00  &42.86 &28.89 &53.33 &45.68 &65.21 \\
Imaginary &63.21  &45.83 &22.22 &43.75 &26.54 &38.29 \\
Shuffled &70.00  &24.17 &12.22 &36.46 &6.12 &8.75 \\
Jumbled &68.75  &30.56 &15.56 &51.04 &46.72 &43.87 \\ \hline
\multicolumn{7}{c}{\textbf{Idefics}} \\ \hline
Original &53.06  &23.96 &4.44 &25.00 &27.66 &61.39 \\
Imaginary &53.06  &0.00 &22.22 &16.67 &0.19 &1.70 \\
Shuffled &51.02  &13.54 &26.67 &21.88 &11.12 &21.09 \\
Jumbled &51.02  &13.54 &8.89 &32.29 &24.07 &47.62 \\ \hline
\multicolumn{7}{c}{\textbf{InternLM}} \\ \hline
Original &53.06  &11.46 &13.33 &17.71 &35.19 &48.64 \\
Imaginary &51.02  &0.00 &17.78 &7.29 &3.32 &7.14 \\
Shuffled &53.06  &12.50 &15.56 &14.58 &8.10 &15.99 \\
Jumbled &34.69  &8.33 &8.89 &8.33 &23.38 &37.59 \\ \hline
\multicolumn{7}{c}{\textbf{CogAgent}} \\ \hline
Original &44.90  &25.00 &15.56 &15.63 &20.48 &38.44 \\
Imaginary &42.86  &0.00 &13.33 &12.50 &2.70 &3.74 \\
Shuffled &51.02  &20.83 &22.22 &23.96 &10.44 &12.59 \\
Jumbled &40.82  &21.88 &17.78 &21.88 &15.27 &29.93 \\ \hline
\multicolumn{7}{c}{\textbf{QwenVL}} \\ \hline
Original &46.94  &14.58 &13.33 &4.17 &13.33 &26.87 \\
Imaginary &51.02  &0.00 &8.89 &6.25 &1.08 &5.61 \\
Shuffled &53.06  &9.38 &17.78 &6.25 &5.58 &13.61 \\
Jumbled &46.94  &18.75 &17.78 &15.63 &13.39 &23.81 \\ \hline
\end{tabular}%
}
\caption{Counter Factual Results (in \%) for EER prompt for all models in the study. CF represents Counter Factual and Acc. stands for Accuracy.}
\label{tab: counterFactual_eer_allresults} 
\end{table}

% Open-source models CogAgent and QwenVL performed worse then there other open-source counterparts specially for Count, Single, Range and List type answers. Interestingly, for binary type answer there performance is on par with other models, which indicates that they are able to identify, extract and compare the value but really struggle with reasoning.

\section{Dataset Validation Process}
\label{sec:dataset_validation_prc}

The ground truth answers were established through a rigorous process: an initial annotation was followed by verification from two additional annotators to ensure accuracy and minimize subjectivity. For region-based question, we adhered to widely accepted geographical definitions and cross referenced them with readily available online resources. 

A total of six annotators were involved in this process. Initial annotation took approximately one minute on average, with more time required for questions involving spatial reasoning or external knowledge about the geographic regions of a country. The verification process was less time consuuming, with each question taking around 20 to 30 seconds on average.

\section{Rank Wise Precision (RWP) Vs MAP and MRR}
\label{sec:rwp_explanation}

\begin{table*}[ht!]\centering
\small
\begin{tabular}{c|cccccc}
\hline

\multirow{2}{*}{\textbf{Map Type}} &  & \textbf{India} &  &  & \textbf{China} &  \\ 
& \textbf{MRR} & \textbf{MAP} & \textbf{RWP} & \textbf{MRR} & \textbf{MAP} & \textbf{RWP} \\ \hline

\toprule
\multicolumn{7}{c}{\textbf{GPT 4o}} \\ \hline
With Annotations &57.41\% &54.94\% &53.70\% &48.40\% &45.66\% &46.58\% \\
Without Annotations &57.41\% &55.25\% &55.56\% &51.21\% &50.52\% &50.48\% \\
Hactched &53.57\% &52.98\% &53.57\% &39.58\% &39.58\% &39.58\% \\ \hline
\multicolumn{7}{c}{\textbf{Gemini 1.5 Flash}} \\ \hline
With Annotations &40.48\% &38.99\% &38.69\% &57.80\% &54.31\% &53.63\% \\
Without Annotations &49.40\% &46.73\% &45.83\% &42.95\% &41.35\% &41.03\% \\
Hactched &38.89\% &38.33\% &38.89\% &61.46\% &60.94\% &60.42\% \\ \hline
\multicolumn{7}{c}{\textbf{Idefics}} \\ \hline
With Annotations &45.24\% &45.24\% &45.24\% &46.15\% &46.15\% &46.15\% \\
Without Annotations &48.81\% &48.81\% &48.81\% &49.36\% &49.36\% &49.36\% \\
Hactched &31.11\% &31.11\% &31.11\% &34.38\% &34.38\% &34.38\% \\ \hline
\multicolumn{7}{c}{\textbf{InternLM}} \\ \hline
With Annotations &35.71\% &35.71\% &35.71\% &42.31\% &41.99\% &41.67\% \\
Without Annotations &39.29\% &39.29\% &39.29\% &39.74\% &39.74\% &39.74\% \\
Hactched &44.44\% &44.44\% &44.44\% &47.92\% &47.92\% &47.92\% \\
\bottomrule

\end{tabular}
\caption{Comparing our RWP score with other popular MRR and MAP rank scores For zero-shot COT prompt}
\label{tab: RWP_MRR_MAP_cot}
\end{table*}

\begin{table*}[ht!]\centering
\small
\begin{tabular}{c|cccccc}
\hline

\multirow{2}{*}{\textbf{Map Type}} &  & \textbf{India} &  &  & \textbf{China} &  \\ 
& \textbf{MRR} & \textbf{MAP} & \textbf{RWP} & \textbf{MRR} & \textbf{MAP} & \textbf{RWP} \\ \hline

\toprule
\multicolumn{7}{c}{\textbf{GPT 4o}} \\ \hline
With Annotations &64.20\% &62.07\% &61.52\% &59.08\% &56.59\% &56.62\% \\
Without Annotations &56.17\% &53.40\% &53.09\% &62.50\% &60.90\% &60.26\% \\
Hactched &64.29\% &63.10\% &61.90\% &39.44\% &38.01\% &38.15\% \\ \hline
\multicolumn{7}{c}{\textbf{Gemini 1.5 Flash}} \\ \hline
With Annotations &38.27\% &36.15\% &35.60\% &53.82\% &50.54\% &49.54\% \\
Without Annotations &61.86\% &60.26\% &60.26\% &41.99\% &37.18\% &35.90\% \\
Hactched &45.24\% &44.05\% &42.86\% &38.54\% &35.94\% &35.42\% \\ \hline
\multicolumn{7}{c}{\textbf{Idefics}} \\ \hline
With Annotations &48.81\% &48.81\% &48.81\% &28.85\% &28.53\% &28.21\% \\
Without Annotations &39.29\% &39.29\% &39.29\% &37.18\% &37.18\% &37.18\% \\
Hactched &26.67\% &26.67\% &26.67\% &39.58\% &39.58\% &39.58\% \\ \hline
\multicolumn{7}{c}{\textbf{InternLM}} \\ \hline
With Annotations &53.57\% &53.57\% &53.57\% &41.67\% &41.67\% &41.67\% \\
Without Annotations &47.62\% &47.62\% &47.62\% &46.15\% &46.15\% &46.15\% \\
Hactched &44.44\% &44.44\% &44.44\% &35.42\% &35.42\% &35.42\% \\
\bottomrule

\end{tabular}
\caption{Comparing our RWP score with other popular MRR and MAP rank scores For EER prompt}
\label{tab: RWP_MRR_MAP_eer}
\end{table*}

The main purpose of introducing the Rank Wise Precision (RWP) score ( Algorithm \ref{alg:RWP_algo} for computing RWP score) was to avoid giving different scores based on the order of the states within the same rank. Traditional metrics such as Mean Reciprocal Rank (MRR) and Mean Average Precision (MAP) assign higher scores to states that appear first in the order. However, for our evaluation, we are concerned with the states irrespective of their order within the same rank. For example, consider the ground truth ranks as follows 
\begin{itemize}
    \item Rank 1: [California]
    \item Rank 2: [Washington]
    \item Rank 3: [Oregon]
\end{itemize}
When the mode is asked to rank the states based on a range value according to the color or shape on a map, it first identifies the color or shape. If more than one state has the same color, they are give the same rank. Consider the two cases:
\begin{itemize}
    \item Case 1: The model's output is Rank 1: [California], Rank 2 : [Washington, Oregon]
    \item Case 2: The model's output is Rank 1: [California], Rank 2: [Oregon, Washington]
\end{itemize}

\begin{algorithm}
\caption{Calculate Rank Wise Precision (RWP) Score}
\label{alg:RWP_algo}
\begin{algorithmic}[1]
\STATE Initialize an empty list $RWP$
\FOR{each rank in ground\_truth\_ranks}
    \STATE $g\_items \gets$ items in the ground\_truth for the current rank
    \STATE $p\_items \gets$ items in predicted order for the current rank
    \STATE Append $precision(g\_items, p\_items)$ to $RWP$
\ENDFOR
\STATE \textbf{return} mean($RWP$)
\end{algorithmic}
\end{algorithm}

In both cases, all three metrics will give a score of 1 for Rank 1 and a score of 0 for Rank 3. However, for Rank 2, MRR will give a score of 1 for Case 1 and 0.5 for Case 2. MAP will give a score of 0.75 for Case 1 and 0.25 for Case 2. In contrast, RWP will give a score of 0.5 for both cases. Therefore, RWP scores are agnostic to the order of states within the same rank, for final score we take the mean of the scores of all 3 ranks. (Table \ref{tab: RWP_MRR_MAP_cot} and \ref{tab: RWP_MRR_MAP_eer} shows the RWP, MAP and MRR scores for India vs China).

\section{Comparison with MapQA dataset}
While MapQA is a valuable resource with its large dataset of 800,000 question-answer pairs, our work distinguishes itself by addressing crucial limitations in MapQA's scope and analytical depth.

\textbf{Targeted Dataset Design and Complexity:}

\begin{itemize}
    \item Our dataset, while smaller in scale than MapQA (3,000 question-answer pairs), is meticulously curated to specifically test complex reasoning skills related to choropleth maps. 
    \item We focus on challenging aspects of choropleth map interpretation, ensuring high-quality data for precise model evaluation.
    \item We incorporate a variety of map types, including continuous and discrete maps with diverse visual representations, such as variations in legend placement, background presence, and colormaps. Additionally, we include real-world map types like hatched maps, increasing the task's complexity.
    \item We analyze both annotated and unannotated maps to further understand how different map types influence question answering performance. 
    \item Unlike MapQA's automatically generated questions, our human-annotated questions require nuanced understanding of relative spatial relationships, intricate map features, and complex reasoning, moving beyond simple information retrieval.
\end{itemize}

For instance, our dataset includes questions such as: “Which two regions that are closest to each other belong to the largest range?” Answering this question necessitates not only identifying the largest range but also using data extraction techniques to find regions within that range. Moreover, models need to rely on visual cues from the map and their internal knowledge base to correctly identify regions that satisfy both the range criteria and proximity requirements.

Another complex example from our dataset is: “Name the southernmost state that belongs to a higher value range compared to all its neighbors.” To answer this, models must extract value data for each state, compare those values with their neighbors, and then utilize visual data or internal knowledge to identify the southernmost state among those meeting the criteria.

\textbf{Additional Diverse Domains:}

\begin{itemize}
    \item MapQA is limited to maps of the USA, whereas our dataset includes maps from three countries (USA, India, and China), helping to highlight potential biases in model understanding of diverse regions.
\end{itemize}

\textbf{Advanced Analysis and Novel Contributions:}

\begin{itemize}
    \item Our analysis surpasses MapQA's scope by encompassing a broader range of models, including open and closed-source Vision-Language Models (VLMs) and Multimodal Language Models (MLLMs). This comprehensive evaluation provides a more accurate picture of the current state-of-the-art in choropleth map understanding and identifies promising avenues for future research.
    \item We go beyond overall accuracy metrics by providing a detailed breakdown of model performance across different answer types. This granular analysis, missing in MapQA, pinpoints areas where models struggle, guiding future research towards targeted improvements in choropleth map understanding.
    \item By evaluating model performance on data with imaginary state names, jumbled state names, and synthetic information, we offer critical insights into model robustness and generalization, pushing the boundaries of current evaluation methods.
\end{itemize}

In conclusion, while MapQA establishes a strong foundation for map-based question answering, our work delves deeper into the complexities of choropleth maps. Our meticulously designed dataset, novel counterfactual analysis, and comprehensive model evaluation provide a more challenging benchmark and a nuanced understanding of model capabilities, paving the way for further advancements in this crucial field.

\section{Zero Shot - CoT Prompt}
\label{sec:zero_shot_cot_prompt_app}
Here is the prompt we used for analysis using zero shot COT.

\begin{figure}[ht!]
      \includegraphics[width=1.0\linewidth]{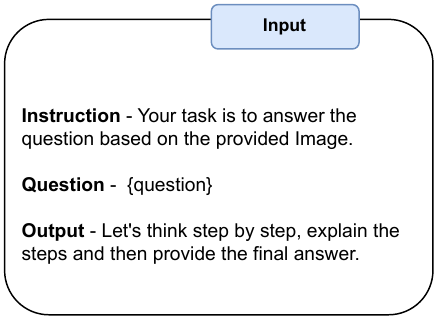}
      \caption{Zero shot COT prompt representation}
      \label{fig:zero_shot_cot_prompt}
\end{figure}

\section{Few Shot - CoT}
\label{sec:few_shot_cot}

\begin{figure}[htb!]
  \includegraphics[width=1.0\linewidth]{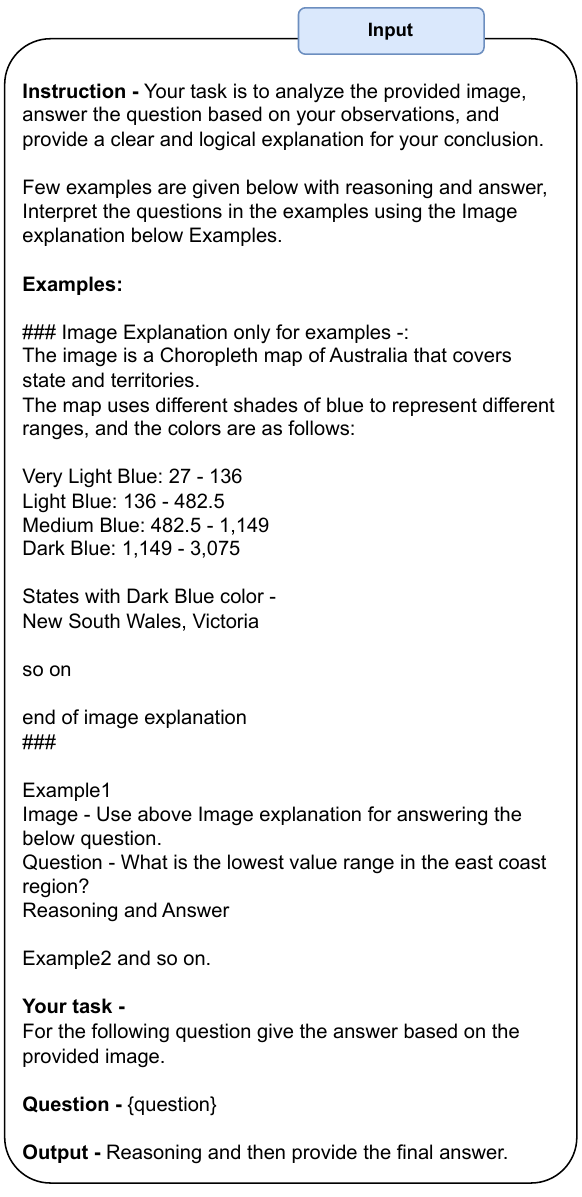}
  \caption{Example of a Few shot COT with visual to textual representation.}
  \label{fig:few_shot_vtm}
\end{figure}

In addition to Zero Shot COT, we also tried Few shot COT. In this approach, we included several examples within the prompt, anticipating that the model would adopt the demonstrated reasoning style before providing its final answer. Given that the task involves both textual and visual modalities, it is crucial to provide different visual cues for the examples to prevent hallucinations caused by manual intervention. We addressed this issue using two sub-approaches:

\begin{table*}[ht!]\centering
\small
\begin{tabular}{c|c|ccccccc}
\hline

\multirow{2}{*}{\textbf{Prompt}} & \multirow{2}{*}{\textbf{Map Type}} & \textbf{Binary} & \textbf{Single} & \textbf{Count} & \textbf{Range} & \textbf{List} & \textbf{List} & \textbf{Rank} \\
&  & \textbf{Acc.} & \textbf{Recall} & \textbf{Acc.} & \textbf{Acc} & \textbf{Prec.} & \textbf{Recall} & \textbf{RWP} \\ \hline

\toprule
\multicolumn{9}{c}{\textbf{USA}} \\ \hline
\multirow{3}{*}{VTM} &With Annotations &68.30\% &61.41\% &55.10\% &66.21\% &38.23\% &46.94\% & -/-\\
&Without Annotations &67.26\% &63.91\% &55.10\% &68.28\% &33.96\% &38.87\% &  -/- \\
&Hatched &54.08\% &57.67\% &51.22\% &48.68\% &27.11\% &40.65\% &  -/-\\ \hline
\multirow{3}{*}{SIE} &With Annotations &64.29\% &50.61\% &46.94\% &59.66\% &31.99\% &37.61\% & -/- \\
&Without Annotations &60.13\% &55.00\% &53.06\% &62.76\% &28.66\% &34.86\% &  -/-\\
&Hatched &50.21\% &56.98\% &40.00\% &47.37\% &30.19\% &37.54\% & -/- \\  \hline
\multicolumn{9}{c}{India} \\ \hline
\multirow{3}{*}{VTM} &With Annotations &65.35\% &45.81\% &23.40\% &45.58\% &36.72\% &42.91\% &38.70\% \\
&Without Annotations &61.62\% &43.80\% &29.79\% &47.35\% &37.28\% &43.75\% &35.63\% \\
&Hatched &57.14\% &48.65\% &29.41\% &40.63\% &30.67\% &33.58\% &40.00\% \\ \hline
\multirow{3}{*}{SIE} &With Annotations &58.33\% &42.31\% &34.04\% &41.59\% &38.03\% &43.06\% &46.55\% \\
&Without Annotations &56.14\% &46.79\% &27.66\% &42.92\% &40.19\% &46.83\% &33.14\% \\
&Hatched &58.79\% &45.27\% &27.45\% &25.00\% &33.63\% &43.31\% &46.67\% \\ \hline
\multicolumn{9}{c}{China} \\ \hline
\multirow{3}{*}{VTM} &With Annotations &60.00\% &56.02\% &14.43\% &52.97\% &29.37\% &36.15\% &41.77\% \\
&Without Annotations &64.55\% &58.60\% &17.53\% &56.78\% &34.56\% &44.79\% &49.47\% \\
&Hatched &52.72\% &47.19\% &29.63\% &38.71\% &30.52\% &44.44\% &29.17\% \\ \hline
\multirow{3}{*}{SIE} &With Annotations &64.77\% &52.26\% &14.43\% &47.03\% &33.10\% &41.92\% &31.84\% \\
&Without Annotations &63.18\% &54.95\% &19.59\% &49.15\% &29.37\% &39.26\% &40.71\% \\
&Hatched &51.63\% &49.57\% &29.63\% &30.65\% &28.43\% &39.32\% &32.81\% \\
\bottomrule

\end{tabular}
\caption{Chain of Thought with Few shot results (in \%) for Gemini model. VTM stands for (visual to textual modality) and SIE stand for (separate image for examples)}
\label{tab: few_shot}
\end{table*}

\begin{figure}[ht!]
  \includegraphics[width=1.0\linewidth]{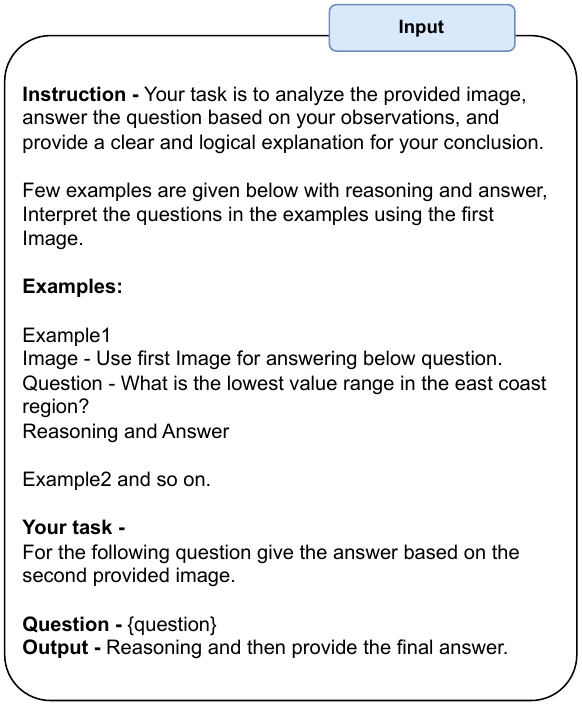}
  \caption{Example of a  Few shot COT with second image for example}
  \label{fig:few_shot_sie}
\end{figure}

\begin{itemize}
    \item \textbf{Textual Conversion of Visual Representation}: The visual map corresponding to example was converted into textual description. (see Figure \ref{fig:few_shot_vtm} for the prompt style)
    \item \textbf{Inclusion of a Second Image in the Prompt}: In this, we provided a separate image for the examples. (see Figure \ref{fig:few_shot_sie} for the prompt style)
\end{itemize}

To avoid introducing any unintended bias through the examples, we prepared examples involving a country not represented in the MAPWise Dataset (Table \ref{tab: few_shot} represents the Few shot results). Largely, Few shots with textual conversion of visual representation (VTM) works better for all map types and country.

% Main Tables Country and Prompt Wise

\section{Comprehensive Results}
\label{sec:comprehensive_results}

In this section, we present the complete results for two prompts - Zero shot COT and Explicit, Extraction and Reasoning (EER) - across all countries, map types and models. This comprehensive coverage provides a detailed comparison of the performance variations under different conditions.

\begin{table*}[ht!]\centering
\small
\begin{tabular}{c|cccccc}
\hline

\multirow{2}{*}{\textbf{Map Type}} & \textbf{Binary} & \textbf{Single} & \textbf{Count} & \textbf{Range} & \textbf{List} & \textbf{List} \\
& \textbf{Acc.} & \textbf{Recall} & \textbf{Acc.} & \textbf{Acc} & \textbf{Prec.} & \textbf{Recall} \\ \hline

\toprule
\multicolumn{7}{c}{\textbf{GPT 4o}} \\ \hline
With Annotations &70.26 &44.68 &51.02 &71.28 &38.64 &45.61 \\
Without Annotations &68.85 &39.38 &53.06 &65.71 &41.58 &47.02 \\
Hactched &49.78 &16.14 &26.83 &26.67 &26.96 &32.60 \\ \hline
\multicolumn{7}{c}{\textbf{Gemini 1.5 Flash}} \\ \hline
With Annotations &65.03 &54.65 &51.02 &63.10 &38.73 &51.43 \\
Without Annotations &65.70 &59.55 &42.86 &64.48 &33.26 &44.04 \\
Hactched &49.36 &56.20 &51.22 &53.95 &20.51 &35.35 \\ \hline
\multicolumn{7}{c}{\textbf{Idefics}} \\ \hline
With Annotations &54.79 &61.11 &18.37 &29.66 &26.64 &42.99 \\
Without Annotations &55.23 &59.09 &24.49 &30.69 &29.61 &43.72 \\
Hactched &52.79 &56.98 &12.20 &25.00 &22.99 &42.09 \\ \hline
\multicolumn{7}{c}{\textbf{InternLM}} \\ \hline
With Annotations &52.78 &35.86 &32.65 &20.00 &22.63 &33.07 \\
Without Annotations &52.78 &42.93 &44.90 &22.07 &20.81 &30.12 \\
Hactched &45.49 &31.40 &53.66 &17.11 &25.84 &31.31 \\ \hline
\multicolumn{7}{c}{\textbf{CogAgent}} \\ \hline
With Annotations &44.03 &42.23 &24.49 &23.40 &19.79 &33.00 \\
Without Annotations &39.34 &42.23 &22.45 &25.18 &20.16 &27.31 \\
Hactched &34.67 &41.34 &24.39 &20.00 &21.66 &25.09 \\ \hline
\multicolumn{7}{c}{\textbf{QwenVL}} \\ \hline
With Annotations &37.72 &20.19 &3.19 &6.82 &19.24 &28.59 \\
Without Annotations &35.09 &16.67 &5.32 &8.18 &18.35 &28.48 \\
Hactched &32.42 &9.46 &3.92 &2.46 &18.42 &24.32 \\
\bottomrule

\end{tabular}
\caption{USA results for all models in the study with zero-shot COT prompt}
\label{tab:usa_cot_main} 
\end{table*}

\begin{table*}[ht]\centering
\small
\begin{tabular}{c|cccccc}
\hline

\multirow{2}{*}{\textbf{Map Type}} & \textbf{Binary} & \textbf{Single} & \textbf{Count} & \textbf{Range} & \textbf{List} & \textbf{List} \\
& \textbf{Acc.} & \textbf{Recall} & \textbf{Acc.} & \textbf{Acc} & \textbf{Prec.} & \textbf{Recall} \\ \hline

\toprule
\multicolumn{7}{c}{\textbf{GPT 4o}} \\ \hline
With Annotations &65.57 &62.02 &51.02 &60.64 &37.78 &45.21 \\
Without Annotations &62.30 &59.60 &59.18 &60.99 &39.00 &44.82 \\
Hactched &36.89 &30.71 &21.95 &28.00 &17.48 &24.36 \\ \hline
\multicolumn{7}{c}{\textbf{Gemini 1.5 Flash}} \\ \hline
With Annotations &66.74 &61.19 &51.02 &62.41 &42.71 &50.28 \\
Without Annotations &64.17 &62.28 &40.82 &61.70 &37.32 &45.05 \\
Hactched &48.44 &40.55 &39.02 &40.00 &18.80 &34.62 \\ \hline
\multicolumn{7}{c}{\textbf{Idefics}} \\ \hline
With Annotations &56.57 &27.02 &6.12 &28.97 &23.80 &46.41 \\
Without Annotations &53.45 &26.26 &12.24 &26.21 &21.58 &45.83 \\
Hactched &56.65 &25.19 &14.63 &19.74 &21.24 &44.87 \\ \hline
\multicolumn{7}{c}{\textbf{InternLM}} \\ \hline
With Annotations &49.67 &22.22 &16.33 &20.69 &24.41 &35.13 \\
Without Annotations &47.66 &33.08 &26.53 &19.66 &25.74 &36.71 \\
Hactched &47.21 &32.17 &31.71 &11.84 &17.72 &26.60 \\ \hline
\multicolumn{7}{c}{\textbf{CogAgent}} \\ \hline
With Annotations &28.10 &20.21 &0.00 &17.38 &19.75 &31.08 \\
Without Annotations &26.70 &30.05 &14.29 &16.67 &20.99 &27.36 \\
Hactched &28.89 &26.77 &9.76 &10.67 &11.15 &15.93 \\ \hline
\multicolumn{7}{c}{\textbf{QwenVL}} \\ \hline
With Annotations &23.46 &16.67 &6.38 &5.45 &19.88 &32.38 \\
Without Annotations &29.82 &9.62 &3.19 &9.09 &21.28 &30.22 \\
Hactched &29.67 &14.19 &5.88 &3.28 &19.94 &26.18 \\
\bottomrule

\end{tabular}
\caption{USA results for all models in the study with EER prompt}
\label{tab:usa_eer_main} 
\end{table*}

\begin{table*}[ht]\centering
\small
\begin{tabular}{c|ccccccc}
\hline

\multirow{2}{*}{\textbf{Map Type}} & \textbf{Binary} & \textbf{Single} & \textbf{Count} & \textbf{Range} & \textbf{List} & \textbf{List} & \textbf{Rank} \\
& \textbf{Acc.} & \textbf{Recall} & \textbf{Acc.} & \textbf{Acc} & \textbf{Prec.} & \textbf{Recall} & \textbf{RWP}\\ \hline

\toprule
\multicolumn{8}{c}{\textbf{GPT 4o}} \\ \hline
With Annotations &71.52 &40.06 &35.48 &55.75 &49.94 &49.17 &54.94 \\
Without Annotations &66.45 &40.92 &30.85 &53.54 &46.23 &47.09 &55.25 \\
Hactched &49.72 &28.38 &31.37 &30.77 &34.19 &36.27 &52.98 \\ \hline
\multicolumn{8}{c}{\textbf{Gemini 1.5 Flash}} \\ \hline
With Annotations &56.36 &38.49 &24.47 &40.27 &34.55 &45.11 &38.99 \\ 
Without Annotations &58.99 &37.39 &23.40 &35.84 &36.25 &46.21 &46.73 \\
Hactched &52.75 &48.65 &23.53 &34.38 &38.95 &47.33 &38.33 \\ \hline
\multicolumn{8}{c}{\textbf{Idefics}} \\ \hline
With Annotations &54.17 &43.59 &13.83 &28.76 &32.19 &38.29 &45.24 \\
Without Annotations &56.58 &43.48 &15.96 &23.45 &28.04 &36.26 &48.81 \\
Hactched &47.80 &45.95 &17.65 &20.31 &23.31 &30.64 &31.11 \\ \hline
\multicolumn{8}{c}{\textbf{InternLM}} \\ \hline
With Annotations &56.80 &32.37 &17.02 &13.27 &20.14 &24.02 &35.71 \\
Without Annotations &53.51 &34.08 &14.89 &13.27 &27.84 &35.84 &39.29 \\
Hactched &51.65 &24.32 &21.57 &24.74 &29.83 &44.44 &44.44 \\ \hline
\multicolumn{8}{c}{\textbf{CogAgent}} \\ \hline
With Annotations &43.27 &25.32 &9.57 &16.81 &19.62 &26.32 &41.36 \\
Without Annotations &44.37 &29.70 &10.64 &15.49 &22.64 &28.86 &38.89 \\
Hactched &43.09 &29.50 &7.84 &4.69 &17.30 &20.32 &46.69 \\ \hline
\multicolumn{8}{c}{\textbf{QwenVL}} \\ \hline
With Annotations &37.75 &22.33 &4.26 &6.64 &17.00 &23.60 &17.31 \\
Without Annotations &35.10 &16.67 &5.32 &8.85 &15.50 &19.55 &35.19 \\
Hactched &32.04 &9.46 &5.88 &2.34 &19.09 &22.55 &21.43 \\
\bottomrule

\end{tabular}
\caption{India results for all models in the study with zero-shot COT prompt}
\label{tab:india_cot_main} 
\end{table*}

\begin{table*}[ht]\centering
\small
\begin{tabular}{c|ccccccc}
\hline

\multirow{2}{*}{\textbf{Map Type}} & \textbf{Binary} & \textbf{Single} & \textbf{Count} & \textbf{Range} & \textbf{List} & \textbf{List} & \textbf{Rank} \\
& \textbf{Acc.} & \textbf{Recall} & \textbf{Acc.} & \textbf{Acc} & \textbf{Prec.} & \textbf{Recall} & \textbf{RWP}\\ \hline

\toprule
\multicolumn{8}{c}{\textbf{GPT 4o}} \\ \hline
With Annotations &65.12 &52.46 &40.00 &52.23 &49.88 &51.51 &62.07 \\
Without Annotations &65.34 &51.82 &40.43 &45.13 &46.98 &47.30 &53.40 \\
Hactched &43.65 &39.86 &33.33 &28.13 &29.46 &30.76 &63.10 \\ \hline
\multicolumn{8}{c}{\textbf{Gemini 1.5 Flash}} \\ \hline
With Annotations &61.37 &37.82 &23.40 &39.38 &29.01 &37.32 &36.15 \\
Without Annotations &62.69 &42.52 &24.47 &43.81 &39.13 &51.44 &60.26 \\
Hactched &58.01 &40.32 &33.33 &31.25 &38.01 &53.24 &44.05 \\ \hline
\multicolumn{8}{c}{\textbf{Idefics}} \\  \hline
With Annotations &55.26 &36.54 &8.51 &21.24 &32.65 &48.75 &48.81 \\
Without Annotations &52.41 &38.25 &11.70 &21.68 &28.17 &46.45 &39.29 \\
Hactched &54.40 &29.73 &5.88 &12.50 &26.31 &37.28 &26.67 \\ \hline
\multicolumn{8}{c}{\textbf{InternLM}} \\ \hline
With Annotations &51.32 &25.96 &15.96 &13.72 &24.20 &27.80 &53.57 \\
Without Annotations &52.41 &28.85 &8.51 &13.72 &25.83 &26.68 &47.62 \\
Hactched &48.35 &13.51 &17.65 &1.56 &17.70 &17.28 &44.44 \\ \hline
\multicolumn{8}{c}{\textbf{CogAgent}} \\ \hline
With Annotations &30.46 &10.90 &4.26 &8.41 &16.55 &19.88 &40.12 \\
Without Annotations &33.33 &13.46 &4.26 &9.29 &17.52 &19.84 &42.42 \\
Hactched &31.49 &9.46 &0.00 &6.25 &15.51 &17.84 &29.23 \\ \hline
\multicolumn{8}{c}{\textbf{QwenVL}} \\ \hline
With Annotations &23.18 &17.95 &5.32 &5.31 &11.64 &20.43 &29.01 \\
Without Annotations &30.02 &9.62 &1.06 &8.85 &16.61 &23.10 &32.10 \\
Hactched &29.83 &14.19 &3.92 &3.13 &17.13 &21.81 &16.67 \\ 
\bottomrule

\end{tabular}
\caption{India results for all models in the study with EER prompt}
\label{tab:india_eer_main} 
\end{table*}

\begin{table*}[ht]\centering
\small
\begin{tabular}{c|ccccccc}
\hline

\multirow{2}{*}{\textbf{Map Type}} & \textbf{Binary} & \textbf{Single} & \textbf{Count} & \textbf{Range} & \textbf{List} & \textbf{List} & \textbf{Rank} \\
& \textbf{Acc.} & \textbf{Recall} & \textbf{Acc.} & \textbf{Acc} & \textbf{Prec.} & \textbf{Recall} & \textbf{RWP}\\ \hline

\toprule
\multicolumn{8}{c}{\textbf{GPT 4o}} \\ \hline
With Annotations &66.97 &47.53 &50.52 &59.40 &53.93 &57.56 &45.66 \\
Without Annotations &68.10 &41.61 &43.18 &61.27 &44.17 &47.32 &50.52 \\
Hactched &45.11 &20.56 &27.78 &18.03 &33.33 &34.40 &39.58 \\ \hline
\multicolumn{8}{c}{\textbf{Gemini 1.5 Flash}} \\ \hline
With Annotations &62.27 &51.83 &13.40 &52.97 &22.76 &38.96 &54.31 \\
Without Annotations &60.23 &51.83 &13.40 &53.81 &31.56 &44.84 &41.35 \\
Hactched &53.80 &55.41 &22.22 &40.32 &29.61 &45.41 &60.94 \\ \hline
\multicolumn{8}{c}{\textbf{Idefics}} \\ \hline
With Annotations &54.09 &50.54 &21.65 &34.32 &21.67 &29.91 &46.15 \\
Without Annotations &56.82 &48.17 &19.59 &30.93 &21.86 &28.73 &49.36 \\
Hactched &56.52 &47.62 &16.67 &14.52 &23.01 &33.33 &34.38 \\  \hline
\multicolumn{8}{c}{\textbf{InternLM}} \\ \hline
With Annotations &54.09 &38.39 &19.59 &28.81 &22.98 &28.02 &41.99 \\
Without Annotations &53.86 &38.49 &22.68 &26.27 &27.28 &33.84 &39.74 \\
Hactched &50.54 &32.90 &22.22 &4.84 &27.37 &31.62 &47.92 \\ \hline
\multicolumn{8}{c}{\textbf{CogAgent}} \\ \hline
With Annotations &48.64 &28.39 &20.62 &22.03 &21.32 &27.91 &48.08 \\
Without Annotations &46.47 &28.60 &11.34 &30.34 &29.08 &31.83 &26.92 \\
Hactched &47.28 &42.64 &11.11 &9.84 &26.39 &32.05 &31.48 \\ \hline
\multicolumn{8}{c}{\textbf{QwenVL}} \\ \hline
With Annotations &37.81 &21.18 &3.09 &9.40 &23.92 &27.90 &21.79 \\
Without Annotations &36.90 &23.66 &2.06 &18.38 &22.56 &27.69 &19.23 \\
Hactched &40.76 &17.32 &11.11 &3.28 &23.34 &27.35 &12.50 \\ \hline
\bottomrule

\end{tabular}
\caption{China results for all models in the study with zero-shot COT prompt}
\label{tab:china_cot_main} 
\end{table*}

\begin{table*}[ht]\centering
\small
\begin{tabular}{c|ccccccc}
\hline

\multirow{2}{*}{\textbf{Map Type}} & \textbf{Binary} & \textbf{Single} & \textbf{Count} & \textbf{Range} & \textbf{List} & \textbf{List} & \textbf{Rank} \\
& \textbf{Acc.} & \textbf{Recall} & \textbf{Acc.} & \textbf{Acc} & \textbf{Prec.} & \textbf{Recall} & \textbf{RWP}\\ \hline

\toprule
\multicolumn{8}{c}{\textbf{GPT 4o}} \\ \hline
With Annotations &63.33 &60.65 &45.36 &59.83 &43.47 &46.48 &56.59 \\
Without Annotations &64.01 &58.39 &49.48 &54.27 &47.24 &50.93 &60.90 \\
Hactched &44.02 &31.39 &24.07 &21.31 &33.01 &37.82 &38.01 \\ \hline
\multicolumn{8}{c}{\textbf{Gemini 1.5 Flash}} \\ \hline
With Annotations &61.50 &54.09 &24.74 &52.14 &23.01 &39.54 &50.54 \\
Without Annotations &59.23 &53.23 &28.87 &53.42 &29.06 &41.46 &37.18 \\
Hactched &47.83 &48.92 &25.93 &36.07 &30.84 &50.43 &35.94 \\ \hline
\multicolumn{8}{c}{\textbf{Idefics}} \\ \hline
With Annotations &53.86 &29.25 &18.56 &28.39 &26.63 &39.78 &28.53 \\
Without Annotations &55.00 &29.78 &19.59 &26.69 &25.70 &37.12 &37.18 \\
Hactched &52.17 &18.18 &12.96 &9.68 &34.03 &47.44 &39.58 \\ \hline
\multicolumn{8}{c}{\textbf{InternLM}} \\ \hline
With Annotations &42.50 &23.66 &21.65 &24.15 &20.97 &24.74 &41.67 \\
Without Annotations &45.91 &32.04 &21.65 &19.92 &24.83 &27.86 &46.15 \\
Hactched &48.37 &20.78 &18.52 &2.42 &30.45 &33.65 &35.42 \\ \hline
\multicolumn{8}{c}{\textbf{CogAgent}} \\ \hline
With Annotations &27.56 &19.35 &5.15 &18.38 &22.25 &25.62 &25.46 \\
Without Annotations &24.15 &16.77 &9.28 &18.38 &27.61 &30.12 &32.05 \\
Hactched &33.70 &22.51 &0.00 &8.20 &31.40 &35.58 &14.58 \\ \hline
\multicolumn{8}{c}{\textbf{QwenVL}} \\ \hline
With Annotations &21.18 &8.92 &2.06 &8.12 &24.01 &29.35 &24.36 \\
Without Annotations &24.37 &13.44 &2.06 &8.55 &22.81 &25.98 &16.67 \\
Hactched &26.09 &7.58 &1.85 &6.56 &20.29 &26.71 &14.58 \\
\bottomrule

\end{tabular}
\caption{China results for all models in the study with EER prompt}
\label{tab:china_eer_main} 
\end{table*}

% Continous Results

\begin{table*}[ht]\centering
\small
\begin{tabular}{c|cccccc}
\hline

\multirow{2}{*}{\textbf{Map Type}} & \textbf{Binary} & \textbf{Single} & \textbf{Count} & \textbf{Range} & \textbf{List} & \textbf{List} \\
& \textbf{Acc.} & \textbf{Recall} & \textbf{Acc.} & \textbf{Acc} & \textbf{Prec.} & \textbf{Recall} \\ \hline

\toprule
\multicolumn{7}{c}{\textbf{GPT 4o}} \\ \hline
With Annotations &64.05\% &62.12\% &25.00\% &61.84\% &36.70\% &42.11\% \\ 
Without Annotations &67.97\% &71.21\% &25.00\% &55.26\% &43.96\% &46.78\% \\ \hline
\multicolumn{7}{c}{\textbf{Gemini 1.5 Flash}} \\ \hline
With Annotations &63.03\% &56.52\% &25.00\% &43.75\% &38.84\% &53.39\% \\
Without Annotations &60.61\% &62.32\% &25.00\% &46.25\% &31.37\% &43.22\% \\ \hline
\multicolumn{7}{c}{\textbf{Idefics}} \\ \hline
With Annotations &61.21\% &63.77\% &12.50\% &27.50\% &23.55\% &41.24\% \\
Without Annotations &61.82\% &60.87\% &50.00\% &36.25\% &29.84\% &41.81\% \\ \hline
\multicolumn{7}{c}{\textbf{InternLM}} \\ \hline
With Annotations &54.55\% &50.72\% &12.50\% &22.50\% &23.56\% &34.18\% \\
Without Annotations &53.94\% &59.42\% &37.50\% &25.00\% &22.40\% &33.90\% \\ \hline
\multicolumn{7}{c}{\textbf{CogAgent}} \\ \hline
With Annotations &44.44\% &16.67\% &12.50\% &31.58\% &20.69\% &30.41\% \\
Without Annotations &39.22\% &13.64\% &12.50\% &27.63\% &25.00\% &32.46\% \\ \hline
\multicolumn{7}{c}{\textbf{QwenVL}} \\ \hline
With Annotations &35.71\% &24.39\% &2.33\% &8.33\% &19.56\% &29.94\% \\
Without Annotations &33.33\% &19.51\% &4.65\% &15.28\% &20.59\% &31.36\% \\
\bottomrule

\end{tabular}
\caption{USA results for all models in the study with zero-shot COT prompt For continuous maps only}
\label{tab:usa_cot_continuos} 
\end{table*}

\begin{table*}[ht]\centering
\small
\begin{tabular}{c|cccccc}
\hline

\multirow{2}{*}{\textbf{Map Type}} & \textbf{Binary} & \textbf{Single} & \textbf{Count} & \textbf{Range} & \textbf{List} & \textbf{List} \\
& \textbf{Acc.} & \textbf{Recall} & \textbf{Acc.} & \textbf{Acc} & \textbf{Prec.} & \textbf{Recall} \\ \hline

\toprule
\multicolumn{7}{c}{\textbf{GPT 4o}} \\ \hline
With Annotations &64.71\% &69.70\% &12.50\% &50.00\% &42.12\% &44.74\% \\
Without Annotations &64.05\% &66.67\% &50.00\% &47.37\% &47.25\% &48.83\% \\ \hline
\multicolumn{7}{c}{\textbf{Gemini 1.5 Flash}} \\ \hline
With Annotations &62.09\% &72.73\% &12.50\% &44.74\% &38.96\% &46.78\% \\
Without Annotations &64.71\% &69.70\% &25.00\% &44.74\% &36.55\% &44.44\% \\ \hline
\multicolumn{7}{c}{\textbf{Idefics}} \\ \hline
With Annotations &60.61\% &20.29\% &12.50\% &35.00\% &27.78\% &48.02\% \\
Without Annotations &59.39\% &20.29\% &25.00\% &32.50\% &24.56\% &46.61\% \\ \hline
\multicolumn{7}{c}{\textbf{InternLM}} \\ \hline
With Annotations &53.94\% &24.64\% &12.50\% &22.50\% &28.63\% &39.55\% \\
Without Annotations &51.52\% &33.33\% &12.50\% &26.25\% &32.94\% &44.07\% \\ \hline
\multicolumn{7}{c}{\textbf{CogAgent}} \\ \hline
With Annotations &28.10\% &27.27\% &0.00\% &22.37\% &24.93\% &40.06\% \\
Without Annotations &25.49\% &16.67\% &0.00\% &19.74\% &26.80\% &33.04\% \\ \hline
\multicolumn{7}{c}{\textbf{QwenVL}} \\ \hline
With Annotations &22.22\% &14.63\% &6.98\% &8.33\% &21.51\% &35.59\% \\
Without Annotations &29.76\% &10.98\% &2.33\% &16.67\% &20.54\% &33.05\% \\
\bottomrule

\end{tabular}
\caption{USA results for all models in the study with EER prompt For continuous maps only}
\label{tab:usa_eer_continuos} 
\end{table*}

\begin{table*}[ht]\centering
\small
\begin{tabular}{c|ccccccc}
\hline

\multirow{2}{*}{\textbf{Map Type}} & \textbf{Binary} & \textbf{Single} & \textbf{Count} & \textbf{Range} & \textbf{List} & \textbf{List} & \textbf{Rank} \\
& \textbf{Acc.} & \textbf{Recall} & \textbf{Acc.} & \textbf{Acc} & \textbf{Prec.} & \textbf{Recall} & \textbf{RWP}\\ \hline

\toprule
\multicolumn{8}{c}{\textbf{GPT 4o}} \\ \hline
With Annotations &72.00\% &40.24\% &26.19\% &58.33\% &54.07\% &53.29\% &65.38\% \\
Without Annotations &70.00\% &41.46\% &25.58\% &51.39\% &52.42\% &54.47\% &71.79\% \\ \hline
\multicolumn{8}{c}{\textbf{Gemini 1.5 Flash}} \\ \hline
With Annotations &53.57\% &29.63\% &18.60\% &43.06\% &37.12\% &51.16\% &58.97\% \\
Without Annotations &59.13\% &28.05\% &23.26\% &43.06\% &37.86\% &49.60\% &74.36\% \\ \hline
\multicolumn{8}{c}{\textbf{Idefics}} \\ \hline
With Annotations &53.97\% &51.22\% &9.30\% &43.06\% &31.53\% &39.38\% &52.56\% \\
Without Annotations &56.35\% &47.56\% &13.95\% &29.17\% &31.70\% &42.39\% &56.41\% \\ \hline
\multicolumn{8}{c}{\textbf{InternLM}} \\ \hline
With Annotations &60.32\% &43.90\% &18.60\% &22.22\% &17.36\% &22.42\% &38.46\% \\
Without Annotations &54.37\% &43.90\% &13.95\% &22.22\% &22.83\% &34.42\% &43.59\% \\ \hline
\multicolumn{8}{c}{\textbf{CogAgent}} \\ \hline
With Annotations &44.40\% &14.63\% &6.98\% &27.78\% &18.06\% &24.29\% &39.74\% \\
Without Annotations &44.00\% &24.39\% &9.30\% &23.61\% &17.98\% &25.71\% &34.62\% \\ \hline
\multicolumn{8}{c}{\textbf{QwenVL}} \\ \hline
With Annotations &35.60\% &25.61\% &4.65\% &8.33\% &17.41\% &24.49\% &32.05\% \\
Without Annotations &33.60\% &19.51\% &4.65\% &15.28\% &15.85\% &20.97\% &32.05\% \\
\bottomrule

\end{tabular}
\caption{India results for all models in the study with zero-shot COT prompt For continuous maps only}
\label{tab:india_cot_continuous} 
\end{table*}

\begin{table*}[ht]\centering
\small
\begin{tabular}{c|ccccccc}
\hline

\multirow{2}{*}{\textbf{Map Type}} & \textbf{Binary} & \textbf{Single} & \textbf{Count} & \textbf{Range} & \textbf{List} & \textbf{List} & \textbf{Rank} \\
& \textbf{Acc.} & \textbf{Recall} & \textbf{Acc.} & \textbf{Acc} & \textbf{Prec.} & \textbf{Recall} & \textbf{RWP}\\ \hline

\toprule
\multicolumn{8}{c}{\textbf{GPT 4o}} \\ \hline
With Annotations &65.60\% &56.10\% &30.23\% &54.17\% &56.10\% &56.40\% &83.33\% \\
Without Annotations &70.40\% &56.10\% &34.88\% &50.00\% &48.63\% &49.90\% &53.85\% \\ \hline
\multicolumn{8}{c}{\textbf{Gemini 1.5 Flash}} \\ \hline
With Annotations &58.00\% &36.59\% &13.95\% &51.39\% &33.29\% &41.46\% &43.16\% \\
Without Annotations &63.20\% &40.24\% &16.28\% &48.61\% &43.51\% &56.81\% &83.33\% \\ \hline
\multicolumn{8}{c}{\textbf{Idefics}} \\ \hline
With Annotations &61.90\% &42.68\% &11.63\% &27.78\% &33.14\% &51.85\% &56.41\% \\
Without Annotations &58.33\% &45.12\% &11.63\% &37.50\% &28.99\% &50.07\% &35.90\% \\ \hline
\multicolumn{8}{c}{\textbf{InternLM}} \\ \hline
With Annotations &52.38\% &32.93\% &18.60\% &26.39\% &27.30\% &32.54\% &50.00\% \\
Without Annotations &51.98\% &35.37\% &4.65\% &23.61\% &25.04\% &28.47\% &44.87\% \\ \hline
\multicolumn{8}{c}{\textbf{CogAgent}} \\ \hline
With Annotations &31.60\% &9.76\% &4.65\% &15.28\% &18.73\% &21.24\% &44.87\% \\ 
Without Annotations &34.40\% &9.76\% &4.65\% &15.28\% &17.78\% &20.26\% &37.18\% \\ \hline
\multicolumn{8}{c}{\textbf{QwenVL}} \\ \hline
With Annotations &21.60\% &15.85\% &6.98\% &8.33\% &10.62\% &18.90\% &30.77\% \\
Without Annotations &30.00\% &10.98\% &2.33\% &16.67\% &13.04\% &20.46\% &28.21\% \\
\bottomrule

\end{tabular}
\caption{India results for all models in the study with EER prompt For continuous maps only}
\label{tab:india_eer_continuous} 
\end{table*}

\begin{table*}[ht]\centering
\small
\begin{tabular}{c|ccccccc}
\hline

\multirow{2}{*}{\textbf{Map Type}} & \textbf{Binary} & \textbf{Single} & \textbf{Count} & \textbf{Range} & \textbf{List} & \textbf{List} & \textbf{Rank} \\
& \textbf{Acc.} & \textbf{Recall} & \textbf{Acc.} & \textbf{Acc} & \textbf{Prec.} & \textbf{Recall} & \textbf{RWP}\\ \hline

\toprule
\multicolumn{8}{c}{\textbf{GPT 4o}} \\ \hline
With Annotations &70.93\% &54.49\% &51.16\% &75.00\% &56.93\% &60.84\% &52.22\% \\
Without Annotations &66.96\% &41.03\% &39.53\% &79.35\% &46.39\% &49.50\% &64.44\% \\ \hline
\multicolumn{8}{c}{\textbf{Gemini 1.5 Flash}} \\ \hline
With Annotations &62.28\% &53.85\% &16.28\% &66.30\% &23.13\% &39.31\% &38.33\% \\
Without Annotations &62.72\% &55.77\% &11.63\% &70.65\% &31.70\% &43.53\% &45.00\% \\ \hline
\multicolumn{8}{c}{\textbf{Idefics}} \\ \hline
With Annotations &50.44\% &50.00\% &25.58\% &57.61\% &19.80\% &26.08\% &43.33\% \\
Without Annotations &51.75\% &46.15\% &25.58\% &53.26\% &19.82\% &25.69\% &56.67\% \\ \hline
\multicolumn{8}{c}{\textbf{InternLM}} \\ \hline
With Annotations &53.95\% &46.15\% &18.60\% &46.74\% &29.17\% &32.94\% &20.00\% \\
Without Annotations &49.12\% &46.15\% &30.23\% &41.30\% &29.79\% &36.67\% &36.67\% \\ \hline
\multicolumn{8}{c}{\textbf{CogAgent}} \\ \hline
With Annotations &45.61\% &16.67\% &13.95\% &41.30\% &23.79\% &29.41\% &50.00\% \\
Without Annotations &42.73\% &15.38\% &6.98\% &46.74\% &32.56\% &33.13\% &33.33\% \\ \hline
\multicolumn{8}{c}{\textbf{QwenVL}} \\ \hline
With Annotations &34.36\% &21.15\% &2.33\% &15.22\% &24.51\% &29.02\% &23.33\% \\ 
Without Annotations &33.92\% &29.49\% &4.65\% &27.17\% &23.93\% &28.51\% &16.67\% \\
\bottomrule

\end{tabular}
\caption{China results for all models in the study with zero-shot COT prompt For continuous maps only}
\label{tab:china_cot_continuous} 
\end{table*}

\begin{table*}[ht]\centering
\small
\begin{tabular}{c|ccccccc}
\hline

\multirow{2}{*}{\textbf{Map Type}} & \textbf{Binary} & \textbf{Single} & \textbf{Count} & \textbf{Range} & \textbf{List} & \textbf{List} & \textbf{Rank} \\
& \textbf{Acc.} & \textbf{Recall} & \textbf{Acc.} & \textbf{Acc} & \textbf{Prec.} & \textbf{Recall} & \textbf{RWP}\\ \hline

\toprule
\multicolumn{8}{c}{\textbf{GPT 4o}} \\ \hline
With Annotations &69.16\% &68.59\% &41.86\% &75.00\% &45.03\% &47.99\% &52.78\% \\
Without Annotations &70.04\% &67.95\% &51.16\% &77.17\% &50.32\% &54.52\% &61.67\% \\ \hline
\multicolumn{8}{c}{\textbf{Gemini 1.5 Flash}} \\ \hline
With Annotations &60.35\% &62.82\% &18.60\% &71.74\% &24.51\% &39.56\% &41.98\% \\
Without Annotations &57.71\% &64.10\% &20.93\% &70.65\% &27.40\% &40.16\% &35.00\% \\ \hline
\multicolumn{8}{c}{\textbf{Idefics}} \\ \hline
With Annotations &56.14\% &33.33\% &18.60\% &46.74\% &23.81\% &32.16\% &30.00\% \\
Without Annotations &55.70\% &33.33\% &23.26\% &42.39\% &24.52\% &32.16\% &43.33\% \\ \hline
\multicolumn{8}{c}{\textbf{InternLM}} \\ \hline
With Annotations &41.23\% &20.51\% &30.23\% &35.87\% &20.60\% &25.88\% &43.33\% \\
Without Annotations &42.54\% &38.46\% &25.58\% &31.52\% &23.14\% &24.31\% &50.00\% \\ \hline
\multicolumn{8}{c}{\textbf{CogAgent}} \\ \hline
With Annotations &26.87\% &8.97\% &0.00\% &35.87\% &20.32\% &23.69\% &40.00\% \\
Without Annotations &23.35\% &6.41\% &2.33\% &29.35\% &27.71\% &29.32\% &43.33\% \\ \hline
\multicolumn{8}{c}{\textbf{QwenVL}} \\ \hline
With Annotations &18.94\% &6.41\% &2.33\% &16.30\% &26.61\% &28.92\% &33.33\% \\
Without Annotations &22.91\% &8.97\% &2.33\% &17.39\% &23.36\% &25.30\% &13.33\% \\
\bottomrule

\end{tabular}
\caption{China results for all models in the study with EER prompt For continuous maps only}
\label{tab:china_eer_continuous} 
\end{table*}

% Discrete Results

\begin{table*}[ht]\centering
\small
\begin{tabular}{c|cccccc}
\hline

\multirow{2}{*}{\textbf{Map Type}} & \textbf{Binary} & \textbf{Single} & \textbf{Count} & \textbf{Range} & \textbf{List} & \textbf{List} \\
& \textbf{Acc.} & \textbf{Recall} & \textbf{Acc.} & \textbf{Acc} & \textbf{Prec.} & \textbf{Recall} \\ \hline

\toprule
\multicolumn{7}{c}{\textbf{GPT 4o}} \\ \hline
With Annotations &73.72\% &35.62\% &56.10\% &74.76\% &39.85\% &47.80\% \\
Without Annotations &69.34\% &22.83\% &58.54\% &69.61\% &40.09\% &47.16\% \\ \hline
\multicolumn{7}{c}{\textbf{Gemini 1.5 Flash}} \\ \hline
With Annotations &66.20\% &53.64\% &56.10\% &70.48\% &38.66\% &50.26\% \\
Without Annotations &68.66\% &58.06\% &46.34\% &71.43\% &34.38\% &44.53\% \\ \hline
\multicolumn{7}{c}{\textbf{Idefics}} \\ \hline
With Annotations &51.06\% &59.69\% &19.51\% &30.48\% &28.48\% &44.02\% \\
Without Annotations &51.41\% &58.14\% &19.51\% &28.57\% &29.48\% &44.87\% \\ \hline
\multicolumn{7}{c}{\textbf{InternLM}} \\ \hline
With Annotations &51.76\% &27.91\% &36.59\% &19.05\% &22.08\% &32.41\% \\
Without Annotations &52.11\% &34.11\% &46.34\% &20.95\% &19.86\% &27.86\% \\ \hline
\multicolumn{7}{c}{\textbf{CogAgent}} \\ \hline
With Annotations &43.80\% &55.51\% &26.83\% &20.39\% &19.23\% &34.62\% \\
Without Annotations &39.42\% &57.09\% &24.39\% &24.27\% &17.13\% &24.08\% \\ \hline
\multicolumn{7}{c}{\textbf{QwenVL}} \\  \hline
With Annotations &40.20\% &15.54\% &3.92\% &6.08\% &19.05\% &27.78\% \\ 
Without Annotations &37.25\% &13.51\% &5.88\% &4.73\% &17.02\% &26.77\% \\
\bottomrule

\end{tabular}
\caption{USA results for all models in the study with zero-shot COT prompt For discrete maps only}
\label{tab:usa_cot_discrete} 
\end{table*}

\begin{table*}[ht]\centering
\small
\begin{tabular}{c|cccccc}
\hline

\multirow{2}{*}{\textbf{Map Type}} & \textbf{Binary} & \textbf{Single} & \textbf{Count} & \textbf{Range} & \textbf{List} & \textbf{List} \\
& \textbf{Acc.} & \textbf{Recall} & \textbf{Acc.} & \textbf{Acc} & \textbf{Prec.} & \textbf{Recall} \\ \hline

\toprule
\multicolumn{7}{c}{\textbf{GPT 4o}} \\ \hline
With Annotations &66.06\% &58.03\% &58.54\% &64.56\% &35.06\% &45.51\% \\
Without Annotations &61.31\% &55.93\% &60.98\% &66.02\% &33.83\% &42.31\% \\ \hline
\multicolumn{7}{c}{\textbf{Gemini 1.5 Flash}} \\ \hline
With Annotations &69.34\% &55.20\% &58.54\% &68.93\% &45.05\% &52.47\% \\
Without Annotations &63.87\% &58.43\% &43.90\% &67.96\% &37.79\% &45.42\% \\ \hline
\multicolumn{7}{c}{\textbf{Idefics}} \\ \hline
With Annotations &54.23\% &30.62\% &4.88\% &26.67\% &21.43\% &45.45\% \\
Without Annotations &50.00\% &29.46\% &9.76\% &23.81\% &19.80\% &45.37\% \\ \hline
\multicolumn{7}{c}{\textbf{InternLM}} \\ \hline
With Annotations &47.18\% &20.93\% &17.07\% &20.00\% &21.89\% &32.49\% \\
Without Annotations &45.42\% &32.95\% &29.27\% &17.14\% &21.46\% &32.32\% \\ \hline
\multicolumn{7}{c}{\textbf{CogAgent}} \\ \hline
With Annotations &28.10\% &16.54\% &0.00\% &15.53\% &16.50\% &25.46\% \\ 
Without Annotations &27.37\% &37.01\% &17.07\% &15.53\% &17.35\% &23.81\% \\ \hline
\multicolumn{7}{c}{\textbf{QwenVL}} \\ \hline
With Annotations &25.00\% &18.92\% &5.88\% &4.05\% &18.90\% &30.47\% \\
Without Annotations &29.90\% &8.11\% &3.92\% &5.41\% &21.71\% &28.54\% \\
\bottomrule

\end{tabular}
\caption{USA results for all models in the study with EER prompt For discrete maps only}
\label{tab:usa_eer_discrete} 
\end{table*}

\begin{table*}[ht]\centering
\small
\begin{tabular}{c|ccccccc}
\hline

\multirow{2}{*}{\textbf{Map Type}} & \textbf{Binary} & \textbf{Single} & \textbf{Count} & \textbf{Range} & \textbf{List} & \textbf{List} & \textbf{Rank} \\
& \textbf{Acc.} & \textbf{Recall} & \textbf{Acc.} & \textbf{Acc} & \textbf{Prec.} & \textbf{Recall} & \textbf{RWP}\\ \hline

\toprule
\multicolumn{8}{c}{\textbf{GPT 4o}} \\ \hline
With Annotations &70.94\% &39.86\% &43.14\% &54.55\% &44.98\% &44.22\% &42.86\% \\
Without Annotations &62.07\% &40.32\% &35.29\% &54.55\% &38.77\% &38.19\% &40.48\% \\ \hline
\multicolumn{8}{c}{\textbf{Gemini 1.5 Flash}} \\ \hline
With Annotations &59.80\% &48.20\% &27.45\% &38.96\% &30.25\% &39.61\% &22.22\% \\
Without Annotations &58.82\% &47.75\% &23.53\% &32.47\% &33.03\% &42.99\% &25.56\% \\ \hline
\multicolumn{8}{c}{\textbf{Idefics}} \\ \hline
With Annotations &54.41\% &35.14\% &17.65\% &22.08\% &33.01\% &36.94\% &35.56\% \\
Without Annotations &56.86\% &38.96\% &17.65\% &20.78\% &23.53\% &28.68\% &44.44\% \\ \hline
\multicolumn{8}{c}{\textbf{InternLM}} \\ \hline
With Annotations &52.45\% &19.59\% &15.69\% &9.09\% &21.62\% &24.29\% &40.00\% \\
Without Annotations &52.45\% &23.20\% &13.73\% &9.09\% &32.07\% &38.28\% &40.00\% \\ \hline
\multicolumn{8}{c}{\textbf{CogAgent}} \\ \hline
With Annotations &41.87\% &37.16\% &11.76\% &11.69\% &21.50\% &28.77\% &42.86\% \\
Without Annotations &44.83\% &35.59\% &11.76\% &11.69\% &28.27\% &32.65\% &42.86\% \\ \hline
\multicolumn{8}{c}{\textbf{QwenVL}} \\ \hline
With Annotations &40.39\% &18.69\% &3.92\% &5.84\% &16.51\% &22.52\% &2.56\% \\
Without Annotations &36.95\% &13.51\% &5.88\% &5.84\% &15.07\% &17.84\% &38.10\% \\
\bottomrule

\end{tabular}
\caption{India results for all models in the study with zero-shot COT prompt For discrete maps only}
\label{tab:india_cot_discrete} 
\end{table*}

\begin{table*}[ht]\centering
\small
\begin{tabular}{c|ccccccc}
\hline

\multirow{2}{*}{\textbf{Map Type}} & \textbf{Binary} & \textbf{Single} & \textbf{Count} & \textbf{Range} & \textbf{List} & \textbf{List} & \textbf{Rank} \\
& \textbf{Acc.} & \textbf{Recall} & \textbf{Acc.} & \textbf{Acc} & \textbf{Prec.} & \textbf{Recall} & \textbf{RWP}\\ \hline

\toprule
\multicolumn{8}{c}{\textbf{GPT 4o}} \\ \hline
With Annotations &64.53\% &48.42\% &48.08\% &51.32\% &42.38\% &45.61\% &41.27\% \\
Without Annotations &59.11\% &47.07\% &45.10\% &42.86\% &45.00\% &44.17\% &52.38\% \\ \hline
\multicolumn{8}{c}{\textbf{Gemini 1.5 Flash}} \\ \hline
With Annotations &65.52\% &39.19\% &31.37\% &33.77\% &23.85\% &32.33\% &28.57\% \\
Without Annotations &62.07\% &45.05\% &31.37\% &41.56\% &33.84\% &44.98\% &40.48\% \\ \hline
\multicolumn{8}{c}{\textbf{Idefics}} \\ \hline
With Annotations &47.06\% &29.73\% &5.88\% &18.18\% &32.03\% &44.93\% &42.22\% \\
Without Annotations &45.10\% &30.63\% &11.76\% &14.29\% &27.16\% &41.99\% &42.22\% \\ \hline
\multicolumn{8}{c}{\textbf{InternLM}} \\ \hline
With Annotations &50.00\% &18.24\% &17.65\% &7.79\% &19.59\% &21.64\% &57.78\% \\
Without Annotations &52.94\% &21.62\% &19.61\% &9.09\% &26.37\% &28.09\% &46.67\% \\ \hline
\multicolumn{8}{c}{\textbf{CogAgent}} \\ \hline
With Annotations &29.06\% &12.16\% &3.92\% &5.19\% &13.92\% &18.24\% &35.71\% \\
Without Annotations &32.02\% &17.57\% &3.92\% &6.49\% &17.21\% &19.34\% &47.22\% \\ \hline
\multicolumn{8}{c}{\textbf{QwenVL}} \\ \hline
With Annotations &25.12\% &20.27\% &3.92\% &3.90\% &12.88\% &22.28\% &30.95\% \\
Without Annotations &30.05\% &8.11\% &0.00\% &5.19\% &20.91\% &26.27\% &35.71\% \\
\bottomrule

\end{tabular}
\caption{India results for all models in the study with EER prompt For discrete maps only}
\label{tab:india_eer_discrete} 
\end{table*}

\begin{table*}[ht]\centering
\small
\begin{tabular}{c|ccccccc}
\hline

\multirow{2}{*}{\textbf{Map Type}} & \textbf{Binary} & \textbf{Single} & \textbf{Count} & \textbf{Range} & \textbf{List} & \textbf{List} & \textbf{Rank} \\
& \textbf{Acc.} & \textbf{Recall} & \textbf{Acc.} & \textbf{Acc} & \textbf{Prec.} & \textbf{Recall} & \textbf{RWP}\\ \hline

\toprule
\multicolumn{8}{c}{\textbf{GPT 4o}} \\ \hline
With Annotations &62.74\% &40.48\% &50.00\% &49.30\% &50.75\% &54.06\% &43.06\% \\
Without Annotations &69.64\% &42.27\% &46.67\% &46.43\% &40.94\% &44.15\% &39.74\% \\ \hline
\multicolumn{8}{c}{\textbf{Gemini 1.5 Flash}} \\ \hline
With Annotations &62.26\% &49.78\% &11.11\% &44.44\% &22.36\% &38.57\% &63.19\% \\
Without Annotations &57.55\% &47.84\% &14.81\% &43.06\% &31.41\% &46.26\% &38.54\% \\ \hline
\multicolumn{8}{c}{\textbf{Idefics}} \\ \hline
With Annotations &58.02\% &51.08\% &18.52\% &19.44\% &23.70\% &34.08\% &47.92\% \\
Without Annotations &62.26\% &50.22\% &14.81\% &16.67\% &24.08\% &32.05\% &44.79\% \\ \hline
\multicolumn{8}{c}{\textbf{InternLM}} \\ \hline
With Annotations &54.25\% &30.52\% &20.37\% &17.36\% &16.24\% &22.65\% &55.21\% \\
Without Annotations &58.96\% &30.74\% &16.67\% &16.67\% &24.55\% &30.77\% &41.67\% \\ \hline
\multicolumn{8}{c}{\textbf{CogAgent}} \\ \hline
With Annotations &47.17\% &38.96\% &18.52\% &15.49\% &24.87\% &27.14\% &29.86\% \\
Without Annotations &50.47\% &41.99\% &14.81\% &19.72\% &25.37\% &30.45\% &22.92\% \\ \hline
\multicolumn{8}{c}{\textbf{QwenVL}} \\ \hline
With Annotations &41.51\% &21.21\% &3.70\% &5.63\% &23.29\% &26.71\% &20.83\% \\
Without Annotations &40.09\% &17.75\% &0.00\% &12.68\% &21.11\% &26.82\% &20.83\% \\
\bottomrule

\end{tabular}
\caption{China results for all models in the study with zero-shot COT prompt For discrete maps only}
\label{tab:china_cot_discrete} 
\end{table*}

\begin{table*}[ht]\centering
\small
\begin{tabular}{c|ccccccc}
\hline

\multirow{2}{*}{\textbf{Map Type}} & \textbf{Binary} & \textbf{Single} & \textbf{Count} & \textbf{Range} & \textbf{List} & \textbf{List} & \textbf{Rank} \\
& \textbf{Acc.} & \textbf{Recall} & \textbf{Acc.} & \textbf{Acc} & \textbf{Prec.} & \textbf{Recall} & \textbf{RWP}\\ \hline

\toprule
\multicolumn{8}{c}{\textbf{GPT 4o}} \\ \hline
With Annotations &57.08\% &52.60\% &48.15\% &50.00\% &41.81\% &44.87\% &59.03\% \\
Without Annotations &57.55\% &48.70\% &48.15\% &39.44\% &43.96\% &47.12\% &59.38\% \\ \hline
\multicolumn{8}{c}{\textbf{Gemini 1.5 Flash}} \\ \hline
With Annotations &62.74\% &45.24\% &29.63\% &39.44\% &21.42\% &39.53\% &54.07\% \\
Without Annotations &60.85\% &42.21\% &35.19\% &42.25\% &30.83\% &42.84\% &36.46\% \\ \hline
\multicolumn{8}{c}{\textbf{Idefics}} \\ \hline
With Annotations &51.42\% &25.11\% &18.52\% &16.67\% &29.70\% &48.08\% &31.25\% \\
Without Annotations &54.25\% &26.19\% &16.67\% &16.67\% &26.99\% &42.52\% &33.33\% \\ \hline
\multicolumn{8}{c}{\textbf{InternLM}} \\ \hline
With Annotations &43.87\% &26.84\% &14.81\% &16.67\% &21.37\% &23.50\% &40.63\% \\
Without Annotations &49.53\% &25.54\% &18.52\% &12.50\% &26.68\% &31.73\% &43.75\% \\ \hline
\multicolumn{8}{c}{\textbf{CogAgent}} \\ \hline
With Annotations &28.30\% &29.87\% &9.26\% &7.04\% &24.30\% &27.67\% &16.32\% \\
Without Annotations &25.00\% &27.27\% &14.81\% &11.27\% &27.49\% &30.98\% &25.00\% \\ \hline
\multicolumn{8}{c}{\textbf{QwenVL}} \\ \hline
With Annotations &23.58\% &11.47\% &1.85\% &2.82\% &21.25\% &29.81\% &18.75\% \\
Without Annotations &25.94\% &17.97\% &1.85\% &2.82\% &22.23\% &26.71\% &22.92\% \\ 
\bottomrule

\end{tabular}
\caption{China results for all models in the study with EER prompt For discrete maps only}
\label{tab:china_eer_discrete} 
\end{table*}

% For relative 

\begin{table*}[ht]\centering
\small
\begin{tabular}{c|cccccc}
\hline

\multirow{2}{*}{\textbf{Map Type}} & \textbf{Binary} & \textbf{Single} & \textbf{Count} & \textbf{Range} & \textbf{List} & \textbf{List} \\
& \textbf{Acc.} & \textbf{Recall} & \textbf{Acc.} & \textbf{Acc} & \textbf{Prec.} & \textbf{Recall} \\ \hline

\toprule
\multicolumn{7}{c}{\textbf{GPT 4o}} \\ \hline
With Annotations &67.77\% &51.56\% &79.17\% &71.28\% &38.64\% &45.61\% \\
Without Annotations &71.08\% &45.94\% &79.17\% &65.71\% &41.58\% &47.02\% \\
Hactched &49.69\% &16.50\% &40.91\% &26.67\% &26.96\% &32.60\% \\ \hline
\multicolumn{7}{c}{\textbf{Gemini 1.5 Flash}} \\ \hline
With Annotations &66.76\% &64.24\% &79.17\% &63.10\% &38.73\% &51.43\% \\
Without Annotations &64.77\% &69.09\% &70.83\% &64.48\% &33.26\% &44.04\% \\
Hactched &51.20\% &63.81\% &81.82\% &53.95\% &20.51\% &35.35\% \\ \hline
\multicolumn{7}{c}{\textbf{Idefics}} \\ \hline
With Annotations &57.67\% &69.09\% &16.67\% &29.66\% &26.64\% &42.99\% \\
Without Annotations &55.97\% &64.85\% &25.00\% &30.69\% &29.61\% &43.72\% \\
Hactched &58.43\% &66.19\% &9.09\% &25.00\% &22.99\% &42.09\% \\ \hline
\multicolumn{7}{c}{\textbf{InternLM}} \\ \hline
With Annotations &56.25\% &36.67\% &45.83\% &20.00\% &22.63\% &33.07\% \\
Without Annotations &55.68\% &42.42\% &58.33\% &22.07\% &20.81\% &30.12\% \\
Hactched &53.01\% &32.86\% &72.73\% &17.11\% &25.84\% &31.31\% \\ \hline
\multicolumn{7}{c}{\textbf{CogAgent}} \\ \hline
With Annotations &45.78\% &45.31\% &41.67\% &23.40\% &19.79\% &33.00\% \\
Without Annotations &40.06\% &47.19\% &37.50\% &25.18\% &20.16\% &27.31\% \\
Hactched &32.70\% &49.03\% &27.27\% &20.00\% &21.66\% &25.09\% \\ \hline
\multicolumn{7}{c}{\textbf{QwenVL}} \\ \hline
With Annotations &36.25\% &21.88\% &4.62\% &6.82\% &19.24\% &28.59\% \\
Without Annotations &33.23\% &18.06\% &7.69\% &8.18\% &18.35\% &28.48\% \\
Hactched &30.89\% &9.68\% &5.41\% &2.46\% &18.42\% &24.32\% \\
\bottomrule

\end{tabular}
\caption{USA results for all models in the study with zero-shot COT prompt For relative questions only}
\label{tab:usa_cot_relative} 
\end{table*}

\begin{table*}[ht]\centering
\small
\begin{tabular}{c|cccccc}
\hline

\multirow{2}{*}{\textbf{Map Type}} & \textbf{Binary} & \textbf{Single} & \textbf{Count} & \textbf{Range} & \textbf{List} & \textbf{List} \\
& \textbf{Acc.} & \textbf{Recall} & \textbf{Acc.} & \textbf{Acc} & \textbf{Prec.} & \textbf{Recall} \\ \hline

\toprule
\multicolumn{7}{c}{\textbf{GPT 4o}} \\ \hline
With Annotations &64.76\% &71.56\% &79.17\% &60.64\% &37.78\% &45.21\% \\
Without Annotations &65.06\% &69.06\% &79.17\% &60.99\% &39.00\% &44.82\% \\
Hactched &38.36\% &36.89\% &40.91\% &28.00\% &17.48\% &24.36\% \\ \hline
\multicolumn{7}{c}{\textbf{Gemini 1.5 Flash}} \\ \hline
With Annotations &68.07\% &67.19\% &75.00\% &62.41\% &42.71\% &50.28\% \\
Without Annotations &64.46\% &69.69\% &70.83\% &61.70\% &37.32\% &45.05\% \\
Hactched &50.94\% &46.12\% &68.18\% &40.00\% &18.80\% &34.62\% \\ \hline
\multicolumn{7}{c}{\textbf{Idefics}} \\ \hline
With Annotations &58.52\% &23.33\% &8.33\% &28.97\% &23.80\% &46.41\% \\
Without Annotations &55.68\% &24.24\% &12.50\% &26.21\% &21.58\% &45.83\% \\
Hactched &59.04\% &23.33\% &18.18\% &19.74\% &21.24\% &44.87\% \\ \hline
\multicolumn{7}{c}{\textbf{InternLM}} \\ \hline
With Annotations &52.84\% &20.00\% &25.00\% &20.69\% &24.41\% &35.13\% \\
Without Annotations &49.15\% &30.61\% &37.50\% &19.66\% &25.74\% &36.71\% \\
Hactched &52.41\% &28.57\% &50.00\% &11.84\% &17.72\% &26.60\% \\ \hline
\multicolumn{7}{c}{\textbf{CogAgent}} \\ \hline
With Annotations &28.31\% &14.37\% &0.00\% &17.38\% &19.75\% &31.08\% \\
Without Annotations &27.11\% &26.56\% &25.00\% &16.67\% &20.99\% &27.36\% \\
Hactched &30.82\% &28.16\% &18.18\% &10.67\% &11.15\% &15.93\% \\ \hline
\multicolumn{7}{c}{\textbf{QwenVL}} \\ \hline
With Annotations &23.87\% &17.36\% &7.69\% &5.45\% &19.88\% &32.38\% \\
Without Annotations &29.91\% &9.72\% &3.08\% &9.09\% &21.28\% &30.22\% \\
Hactched &29.27\% &15.32\% &2.70\% &3.28\% &19.94\% &26.18\% \\
\bottomrule

\end{tabular}
\caption{USA results for all models in the study with EER prompt For relative questions only}
\label{tab:usa_eer_relative} 
\end{table*}

\begin{table*}[ht]\centering
\small
\begin{tabular}{c|ccccccc}
\hline

\multirow{2}{*}{\textbf{Map Type}} & \textbf{Binary} & \textbf{Single} & \textbf{Count} & \textbf{Range} & \textbf{List} & \textbf{List} & \textbf{Rank} \\
& \textbf{Acc.} & \textbf{Recall} & \textbf{Acc.} & \textbf{Acc} & \textbf{Prec.} & \textbf{Recall} & \textbf{RWP}\\ \hline

\toprule
\multicolumn{8}{c}{\textbf{GPT 4o}} \\ \hline
With Annotations &71.43\% &41.67\% &35.38\% &55.75\% &49.94\% &49.17\% &53.70\% \\
Without Annotations &65.65\% &41.67\% &27.69\% &53.54\% &46.23\% &47.09\% &55.56\% \\
Hactched &48.78\% &26.61\% &35.14\% &30.77\% &34.19\% &36.27\% &53.57\% \\ \hline
\multicolumn{8}{c}{\textbf{Gemini 1.5 Flash}} \\ \hline
With Annotations &56.19\% &40.21\% &18.46\% &40.27\% &34.04\% &45.99\% &39.29\% \\
\bottomrule

\end{tabular}
\caption{India results for all models in the study with zero-shot COT prompt For relative questions only}
\label{tab:india_cot_relative} 
\end{table*}

\begin{table*}[ht]\centering
\small
\begin{tabular}{c|ccccccc}
\hline

\multirow{2}{*}{\textbf{Map Type}} & \textbf{Binary} & \textbf{Single} & \textbf{Count} & \textbf{Range} & \textbf{List} & \textbf{List} & \textbf{Rank} \\
& \textbf{Acc.} & \textbf{Recall} & \textbf{Acc.} & \textbf{Acc} & \textbf{Prec.} & \textbf{Recall} & \textbf{RWP}\\ \hline

\toprule
\multicolumn{8}{c}{\textbf{GPT 4o}} \\ \hline
With Annotations &63.83\% &55.21\% &45.45\% &52.23\% &49.88\% &51.51\% &61.52\% \\
Without Annotations &67.17\% &53.47\% &38.46\% &45.13\% &46.98\% &47.30\% &53.09\% \\
Hactched &43.90\% &41.94\% &40.54\% &28.13\% &29.46\% &30.76\% &61.90\% \\ \hline
\multicolumn{8}{c}{\textbf{Gemini 1.5 Flash}} \\ \hline
With Annotations &62.01\% &37.85\% &27.69\% &39.38\% &29.01\% &37.32\% &35.60\% \\
\bottomrule

\end{tabular}
\caption{India results for all models in the study with EER prompt For relative questions only}
\label{tab:india_eer_relative} 
\end{table*}

\begin{table*}[ht]\centering
\small
\begin{tabular}{c|ccccccc}
\hline

\multirow{2}{*}{\textbf{Map Type}} & \textbf{Binary} & \textbf{Single} & \textbf{Count} & \textbf{Range} & \textbf{List} & \textbf{List} & \textbf{Rank} \\
& \textbf{Acc.} & \textbf{Recall} & \textbf{Acc.} & \textbf{Acc} & \textbf{Prec.} & \textbf{Recall} & \textbf{RWP}\\ \hline

\toprule
\multicolumn{8}{c}{\textbf{GPT 4o}} \\ \hline
With Annotations &68.75\% &50.12\% &48.61\% &59.05\% &53.93\% &57.56\% &46.58\% \\
Without Annotations &70.07\% &42.35\% &47.62\% &61.88\% &44.17\% &47.32\% &50.48\% \\
Hactched &40.94\% &24.74\% &31.82\% &18.03\% &33.33\% &34.40\% &39.58\% \\ \hline
\multicolumn{8}{c}{\textbf{Gemini 1.5 Flash}} \\ \hline
With Annotations &61.68\% &54.14\% &15.28\% &53.42\% &22.76\% &38.96\% &53.63\% \\
\bottomrule

\end{tabular}
\caption{China results for all models in the study with zero-shot COT prompt For relative questions only}
\label{tab:china_cot_relative} 
\end{table*}

\begin{table*}[ht]\centering
\small
\begin{tabular}{c|ccccccc}
\hline

\multirow{2}{*}{\textbf{Map Type}} & \textbf{Binary} & \textbf{Single} & \textbf{Count} & \textbf{Range} & \textbf{List} & \textbf{List} & \textbf{Rank} \\
& \textbf{Acc.} & \textbf{Recall} & \textbf{Acc.} & \textbf{Acc} & \textbf{Prec.} & \textbf{Recall} & \textbf{RWP}\\ \hline

\toprule
\multicolumn{8}{c}{\textbf{GPT 4o}} \\ \hline
With Annotations &65.31\% &65.96\% &48.61\% &59.48\% &43.47\% &46.48\% &56.62\% \\
Without Annotations &66.88\% &62.06\% &48.61\% &54.74\% &47.24\% &50.93\% &60.26\% \\
Hactched &40.94\% &37.76\% &27.27\% &21.31\% &33.01\% &37.82\% &38.15\% \\ \hline
\multicolumn{8}{c}{\textbf{Gemini 1.5 Flash}} \\ \hline
With Annotations &65.63\% &57.33\% &29.17\% &52.59\% &23.01\% &39.54\% &49.54\% \\
\bottomrule

\end{tabular}
\caption{China results for all models in the study with EER prompt For relative questions only}
\label{tab:china_eer_relative} 
\end{table*}

% For non_relative

\begin{table*}[ht]\centering
\small
\begin{tabular}{c|cccccc}
\hline

\multirow{2}{*}{\textbf{Map Type}} & \textbf{Binary} & \textbf{Single} & \textbf{Count} & \textbf{Range} & \textbf{List} & \textbf{List} \\
& \textbf{Acc.} & \textbf{Recall} & \textbf{Acc.} & \textbf{Acc} & \textbf{Prec.} & \textbf{Recall} \\ \hline

\toprule
\multicolumn{7}{c}{\textbf{GPT 4o}} \\ \hline
With Annotations &78.95\% &11.31\% &24.00\% &0.00\% &0.00\% &0.00\% \\
Without Annotations &61.05\% &7.58\% &28.00\% &0.00\% &0.00\% &0.00\% \\
Hactched &50.00\% &14.58\% &10.53\% &0.00\% &0.00\% &0.00\% \\ \hline
\multicolumn{7}{c}{\textbf{Gemini 1.5 Flash}} \\ \hline
With Annotations &58.76\% &6.67\% &24.00\% &0.00\% &0.00\% &0.00\% \\
Without Annotations &69.07\% &11.82\% &16.00\% &0.00\% &0.00\% &0.00\% \\
Hactched &44.78\% &22.92\% &15.79\% &0.00\% &0.00\% &0.00\% \\ \hline
\multicolumn{7}{c}{\textbf{Idefics}} \\ \hline
With Annotations &44.33\% &21.21\% &20.00\% &0.00\% &0.00\% &0.00\% \\
Without Annotations &52.58\% &30.30\% &24.00\% &0.00\% &0.00\% &0.00\% \\
Hactched &38.81\% &16.67\% &15.79\% &0.00\% &0.00\% &0.00\% \\ \hline
\multicolumn{7}{c}{\textbf{InternLM}} \\ \hline
With Annotations &40.21\% &31.82\% &20.00\% &0.00\% &0.00\% &0.00\% \\
Without Annotations &42.27\% &45.45\% &32.00\% &0.00\% &0.00\% &0.00\% \\
Hactched &26.87\% &25.00\% &31.58\% &0.00\% &0.00\% &0.00\% \\ \hline
\multicolumn{7}{c}{\textbf{CogAgent}} \\ \hline
With Annotations &37.89\% &27.27\% &8.00\% &0.00\% &0.00\% &0.00\% \\
Without Annotations &36.84\% &18.18\% &8.00\% &0.00\% &0.00\% &0.00\% \\
Hactched &39.39\% &8.33\% &21.05\% &0.00\% &0.00\% &0.00\% \\ \hline
\multicolumn{7}{c}{\textbf{QwenVL}} \\ \hline
With Annotations &41.60\% &0.00\% &0.00\% &0.00\% &0.00\% &0.00\% \\
Without Annotations &40.00\% &0.00\% &0.00\% &0.00\% &0.00\% &0.00\% \\
Hactched &35.59\% &8.33\% &0.00\% &0.00\% &0.00\% &0.00\% \\
\bottomrule

\end{tabular}
\caption{USA results for all models in the study with zero-shot COT prompt For non-relative questions only}
\label{tab:usa_cot_non_relative} 
\end{table*}

\begin{table*}[ht]\centering
\small
\begin{tabular}{c|cccccc}
\hline

\multirow{2}{*}{\textbf{Map Type}} & \textbf{Binary} & \textbf{Single} & \textbf{Count} & \textbf{Range} & \textbf{List} & \textbf{List} \\
& \textbf{Acc.} & \textbf{Recall} & \textbf{Acc.} & \textbf{Acc} & \textbf{Prec.} & \textbf{Recall} \\ \hline

\toprule
\multicolumn{7}{c}{\textbf{GPT 4o}} \\ \hline
With Annotations &68.42\% &15.76\% &24.00\% &0.00\% &0.00\% &0.00\% \\
Without Annotations &52.63\% &13.74\% &40.00\% &0.00\% &0.00\% &0.00\% \\
Hactched &33.33\% &4.17\% &0.00\% &0.00\% &0.00\% &0.00\% \\ \hline
\multicolumn{7}{c}{\textbf{Gemini 1.5 Flash}} \\ \hline
With Annotations &62.11\% &32.12\% &28.00\% &0.00\% &0.00\% &0.00\% \\
Without Annotations &63.16\% &26.36\% &12.00\% &0.00\% &0.00\% &0.00\% \\
Hactched &42.42\% &16.67\% &5.26\% &0.00\% &0.00\% &0.00\% \\ \hline
\multicolumn{7}{c}{\textbf{Idefics}} \\ \hline
With Annotations &49.48\% &45.45\% &4.00\% &0.00\% &0.00\% &0.00\% \\
Without Annotations &45.36\% &36.36\% &12.00\% &0.00\% &0.00\% &0.00\% \\
Hactched &50.75\% &33.33\% &10.53\% &0.00\% &0.00\% &0.00\% \\ \hline
\multicolumn{7}{c}{\textbf{InternLM}} \\ \hline
With Annotations &38.14\% &33.33\% &8.00\% &0.00\% &0.00\% &0.00\% \\
Without Annotations &42.27\% &45.45\% &16.00\% &0.00\% &0.00\% &0.00\% \\
Hactched &34.33\% &47.92\% &10.53\% &0.00\% &0.00\% &0.00\% \\ \hline
\multicolumn{7}{c}{\textbf{CogAgent}} \\ \hline
With Annotations &27.37\% &48.48\% &0.00\% &0.00\% &0.00\% &0.00\% \\
Without Annotations &25.26\% &46.97\% &4.00\% &0.00\% &0.00\% &0.00\% \\
Hactched &24.24\% &20.83\% &0.00\% &0.00\% &0.00\% &0.00\% \\ \hline
\multicolumn{7}{c}{\textbf{QwenVL}} \\ \hline
With Annotations &22.40\% &8.33\% &3.45\% &0.00\% &0.00\% &0.00\% \\
Without Annotations &29.60\% &8.33\% &3.45\% &0.00\% &0.00\% &0.00\% \\
Hactched &30.51\% &8.33\% &14.29\% &0.00\% &0.00\% &0.00\% \\
\bottomrule

\end{tabular}
\caption{USA results for all models in the study with EER prompt For non-relative questions only}
\label{tab:usa_eer_non_relative} 
\end{table*}

\begin{table*}[ht]\centering
\small
\begin{tabular}{c|ccccccc}
\hline

\multirow{2}{*}{\textbf{Map Type}} & \textbf{Binary} & \textbf{Single} & \textbf{Count} & \textbf{Range} & \textbf{List} & \textbf{List} & \textbf{Rank} \\
& \textbf{Acc.} & \textbf{Recall} & \textbf{Acc.} & \textbf{Acc} & \textbf{Prec.} & \textbf{Recall} & \textbf{RWP}\\ \hline

\toprule
\multicolumn{8}{c}{\textbf{GPT 4o}} \\ \hline
With Annotations &71.77\% &20.83\% &35.71\% &0.00\% &0.00\% &0.00\% &0.00\% \\
Without Annotations &68.55\% &31.94\% &37.93\% &0.00\% &0.00\% &0.00\% &0.00\% \\
Hactched &51.72\% &37.50\% &21.43\% &0.00\% &0.00\% &0.00\% &0.00\% \\ \hline
\multicolumn{8}{c}{\textbf{Gemini 1.5 Flash}} \\ \hline
With Annotations &56.80\% &18.06\% &34.48\% &0.00\% &0.00\% &0.00\% &0.00\% \\
\bottomrule

\end{tabular}
\caption{India results for all models in the study with zero-shot COT prompt For non-relative questions only}
\label{tab:india_cot_non_relative} 
\end{table*}

\begin{table*}[ht]\centering
\small
\begin{tabular}{c|ccccccc}
\hline

\multirow{2}{*}{\textbf{Map Type}} & \textbf{Binary} & \textbf{Single} & \textbf{Count} & \textbf{Range} & \textbf{List} & \textbf{List} & \textbf{Rank} \\
& \textbf{Acc.} & \textbf{Recall} & \textbf{Acc.} & \textbf{Acc} & \textbf{Prec.} & \textbf{Recall} & \textbf{RWP}\\ \hline

\toprule
\multicolumn{8}{c}{\textbf{GPT 4o}} \\ \hline
With Annotations &68.55\% &19.44\% &27.59\% &0.00\% &0.00\% &0.00\% &0.00\% \\
Without Annotations &60.48\% &31.94\% &44.83\% &0.00\% &0.00\% &0.00\% &0.00\% \\
Hactched &43.10\% &29.17\% &14.29\% &0.00\% &0.00\% &0.00\% &0.00\% \\ \hline
\multicolumn{8}{c}{\textbf{Gemini 1.5 Flash}} \\ \hline
With Annotations &59.68\% &37.50\% &13.79\% &0.00\% &0.00\% &0.00\% &0.00\% \\
\bottomrule

\end{tabular}
\caption{India results for all models in the study with EER prompt For non-relative questions only}
\label{tab:india_eer_non_relative} 
\end{table*}

\begin{table*}[ht]\centering
\small
\begin{tabular}{c|ccccccc}
\hline

\multirow{2}{*}{\textbf{Map Type}} & \textbf{Binary} & \textbf{Single} & \textbf{Count} & \textbf{Range} & \textbf{List} & \textbf{List} & \textbf{Rank} \\
& \textbf{Acc.} & \textbf{Recall} & \textbf{Acc.} & \textbf{Acc} & \textbf{Prec.} & \textbf{Recall} & \textbf{RWP}\\ \hline

\toprule
\multicolumn{8}{c}{\textbf{GPT 4o}} \\ \hline
With Annotations &62.18\% &21.43\% &56.00\% &100.00\% &0.00\% &0.00\% &0.00\% \\
Without Annotations &63.06\% &33.33\% &32.00\% &0.00\% &0.00\% &0.00\% &0.00\% \\
Hactched &54.39\% &0.00\% &10.00\% &0.00\% &0.00\% &0.00\% &0.00\% \\ \hline
\multicolumn{8}{c}{\textbf{Gemini 1.5 Flash}} \\ \hline
With Annotations &63.87\% &28.57\% &8.00\% &0.00\% &0.00\% &0.00\% &0.00\% \\
\bottomrule

\end{tabular}
\caption{China results for all models in the study with zero-shot COT prompt For non-relative questions only}
\label{tab:china_cot_non_relative} 
\end{table*}

\begin{table*}[ht]\centering
\small
\begin{tabular}{c|ccccccc}
\hline

\multirow{2}{*}{\textbf{Map Type}} & \textbf{Binary} & \textbf{Single} & \textbf{Count} & \textbf{Range} & \textbf{List} & \textbf{List} & \textbf{Rank} \\
& \textbf{Acc.} & \textbf{Recall} & \textbf{Acc.} & \textbf{Acc} & \textbf{Prec.} & \textbf{Recall} & \textbf{RWP}\\ \hline

\toprule
\multicolumn{8}{c}{\textbf{GPT 4o}} \\ \hline
With Annotations &57.98\% &7.14\% &36.00\% &100.00\% &0.00\% &0.00\% &0.00\% \\
Without Annotations &56.30\% &21.43\% &52.00\% &0.00\% &0.00\% &0.00\% &0.00\% \\
Hactched &50.88\% &0.00\% &10.00\% &0.00\% &0.00\% &0.00\% &0.00\% \\ \hline
\multicolumn{8}{c}{\textbf{Gemini 1.5 Flash}} \\ \hline
With Annotations &50.42\% &21.43\% &12.00\% &0.00\% &0.00\% &0.00\% &0.00\% \\
\bottomrule

\end{tabular}
\caption{China results for all models in the study with EER prompt For non-relative questions only}
\label{tab:china_eer_non_relative} 
\end{table*}

\end{document}